\documentclass[runningheads]{llncs}

\usepackage[mobile]{eccv}

\usepackage{eccvabbrv}
\usepackage{tikz}
\usepackage{comment}
\usepackage{amsmath,amssymb} 
\usepackage{color}
\usepackage{gensymb}
\usepackage{makecell}
\usepackage{wrapfig}
\usepackage[noend]{algpseudocode}
\usepackage{algorithmicx,algorithm}
\usepackage{textcomp}
\usepackage{xcolor}
\usepackage{colortbl}
\usepackage{graphicx}
\usepackage{booktabs}
\usepackage{multirow}
\usepackage[accsupp]{axessibility}  

\definecolor{r}{rgb}{1,0,0}
\definecolor{g}{rgb}{0,0.69,0.3137}
\definecolor{yellow}{rgb}{1, 1, 0.7}
\definecolor{orange}{rgb}{1, 0.85, 0.7}
\definecolor{red2}{rgb}{1, 0.7, 0.7}
\usepackage[pagebackref,breaklinks,colorlinks,citecolor=eccvblue]{hyperref}

\usepackage{orcidlink}

\begin{document}

\title{Raising the Ceiling: Conflict-Free Local Feature Matching with Dynamic View Switching} 

\titlerunning{RCM: Conflict-Free Local Feature Matching with Dynamic View Switching}

\author{Xiaoyong Lu \and
Songlin Du\thanks{Corresponding author}}

\authorrunning{X.~Lu et al.}

\institute{School of Automation, Southeast University, Nanjing, China\\
\email{\{luxiaoyong,sdu\}@seu.edu.cn}
}

\maketitle

\begin{abstract}
Current feature matching methods prioritize improving modeling capabilities to better align outputs with ground-truth matches,
which are the theoretical upper bound on matching results, metaphorically depicted as the ``ceiling''.
However, these enhancements fail to address the underlying issues that directly hinder ground-truth matches, 
including the scarcity of matchable points in small scale images,
matching conflicts in dense methods,
and the keypoint-repeatability reliance in sparse methods.
We propose a novel feature matching method named RCM, which \textbf{R}aises the \textbf{C}eiling of \textbf{M}atching from three aspects.
1) RCM introduces a dynamic view switching mechanism to address the scarcity of matchable points in source images by strategically switching image pairs.
2) RCM proposes a conflict-free coarse matching module, addressing matching conflicts in the target image through a many-to-one matching strategy.
3) By integrating the semi-sparse paradigm and the coarse-to-fine architecture, 
RCM preserves the benefits of both high efficiency and global search, mitigating the reliance on keypoint repeatability.
As a result, 
RCM enables more matchable points in the source image to be matched in an exhaustive and conflict-free manner in the target image,
leading to a substantial $260\%$ increase in ground-truth matches.
Comprehensive experiments show that RCM exhibits remarkable performance and efficiency in comparison to state-of-the-art methods.

\end{abstract}
    
\section{Introduction}
\label{sec:intro}

\begin{figure}[t]
	\includegraphics[width=0.99\linewidth]{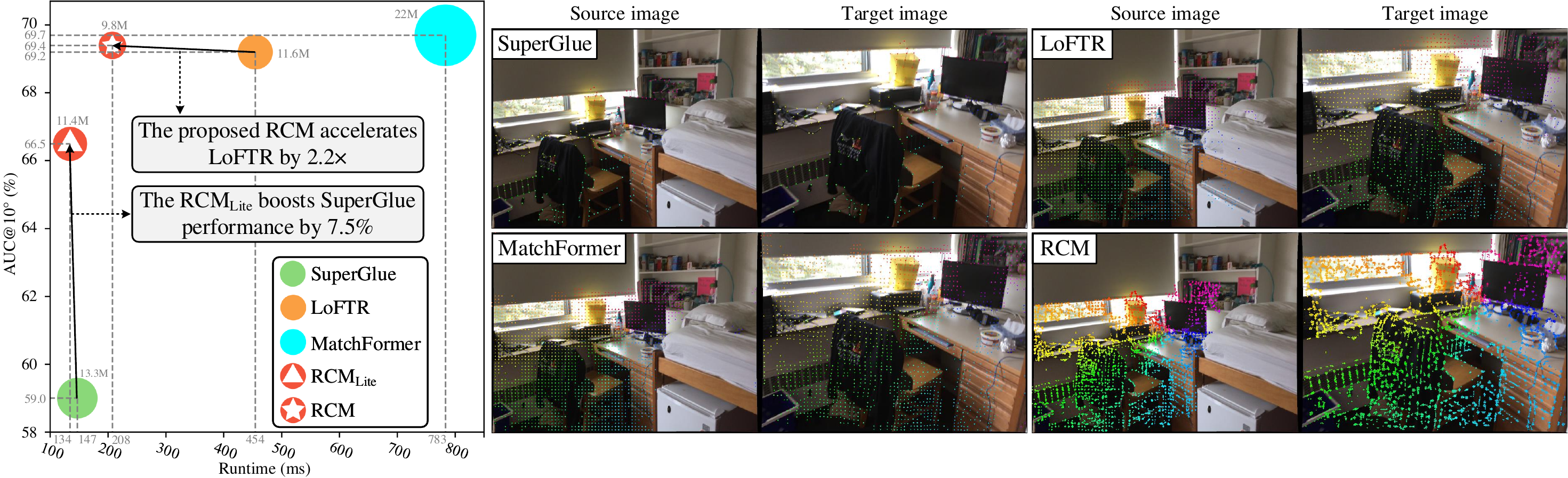}
	\centering
	\caption{\textbf{Comparison among RCM, RCM$_{\mathrm{\mathbf{Lite}}}$, SuperGlue \cite{superglue}, LoFTR \cite{loftr} and MatchFormer \cite{matchformer}.}
	Full RCM significantly accelerates LoFTR by $2.2\times$ while achieving superior performance.
	The lightweight RCM$_{\mathrm{Lite}}$ outperforms SuperGlue by $+7.5\%$ while maintaining a speed advantage.
	The same color indicates the matched features in the visualization.
	}
	 \label{figure1}
\end{figure}

Feature matching, which aims to establish accurate correspondences between two images, 
underpins various pivotal tasks in computer vision, such as object recognition \cite{sift-flow}, 
structure from motion (SfM) \cite{sfm}, and simultaneous localization and mapping (SLAM) \cite{slam}.
However, challenges arise from factors such as viewpoint change, illumination variation, scale variation, occlusion and motion blur between images.
These complexities elevate the difficulty of feature matching, particularly when real-time performance is required for practical applications.

Presently, methods for feature matching can be broadly categorized into two groups: sparse matching methods and dense matching methods.
Sparse methods \cite{superglue, sgmnet, clustergnn, paraformer} are designed to establish correspondences between two sets of keypoints. 
Consequently, the incorporation of detectors \cite{sift, orb, superpoint, r2d2} becomes essential in this process.
In contrast, dense methods \cite{loftr, patch2pix, cotr, ncnet} undertake a direct pursuit of matches between the dense features.

Through a comprehensive examination of prevailing sparse and dense matching methods, 
we identify and subsequently address three key issues that constrain the theoretical upper bound, \ie, the ceiling, on the number of matches.

\textbf{(1) Sparse and dense methods: dramatic decrease in the number of matchable keypoints in small scale images.}
Overlapping regions in small scale images are limited in size, with the majority of keypoints or grid points situated in irrelevant background, 
which not only causes redundancy in the message passing but also radically limits the number of ground-truth matches.

For the semi-sparse \cite{s2dhm,s2dnet} and dense \cite{loftr} matching paradigms, 
this challenge only arises in the source image as they perform a dense search in the target image.
To tackle this challenge, we propose a dynamic view switcher that learns to switch the larger scale image to the source image. 
This strategic switch significantly augments the number of matchable keypoints within the overlapping region to radically raise the theoretical upper bound of the matcher, 
as illustrated in \cref{fig:m2o_switcher_viz}(b).

\begin{figure*}[t]
	\includegraphics[width=0.99\linewidth]{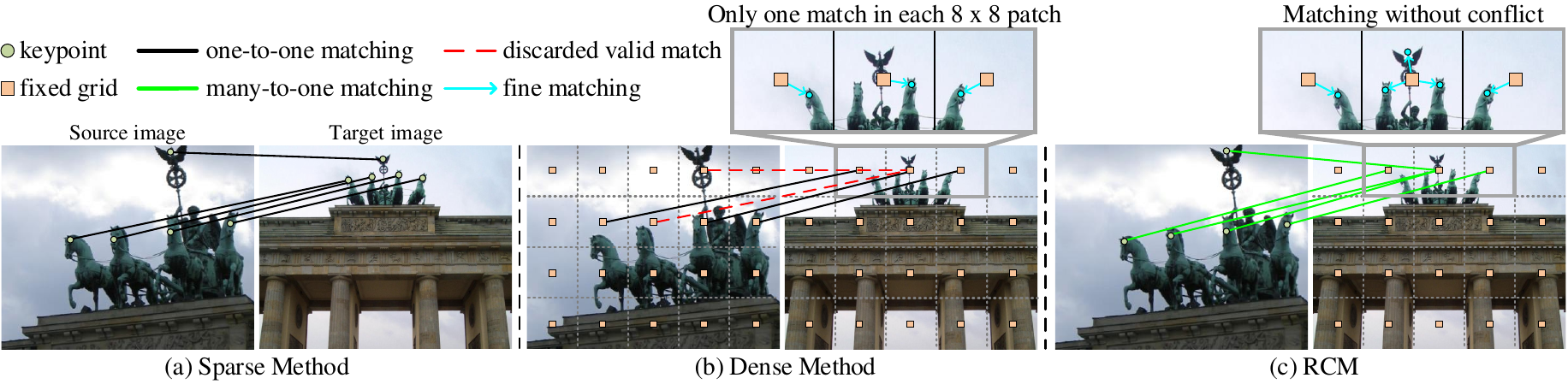}
	\centering
	\caption{\textbf{Comparison of three matching methods.}
	\textbf{(a)} The sparse methods rely on precise keypoint detection at corresponding points in both images.
	\textbf{(b)} We highlight the issues of matching conflicts introduced in \cref{sec:intro}.
	\textbf{(c)} RCM establishes conflict-free coarse matches and adjusts them independently in the fine matching stage.}
	 \label{figure2}
\end{figure*}

\textbf{(2) Dense method: matching conflicts in large scale variation scenes caused by the one-to-one matching strategy.}
As shown in \cref{figure2}(b), the case of many-to-one matching arises when dealing with significant scale variations, 
as more pixels in large scale images correspond to fewer pixels in small scale images.
However, existing semi-dense methods struggle with this challenge as they employ a one-to-one matching strategy in the coarse matching stage, 
generating at most one coarse match within each $8\times8$ region in the target image while discarding other valid matches.

In our method, 
correspondences are established between the source and target image in a many-to-one manner.
Each matchable point can independently search for correspondences without conflicts, as depicted in \cref{figure2}(c) and \cref{fig:m2o_switcher_viz}(a).

\textbf{(3) Sparse method: heavy reliance on the repeatability of keypoint.}
Sparse methods strategically narrow the search scope for matchers by leveraging detectors, 
yet they concurrently establish a severe reliance on the repeatability of keypoints.
When input keypoints are not precisely detected at corresponding positions in both images and result in few ground-truth matches in challenging scenes, 
even a perfect sparse matcher fails to ``create something out of nothing''.

This challenge inspires the semi-sparse paradigm \cite{s2dhm,s2dnet} to detect keypoints in the source image and perform a dense search in the target image, 
eliminating the requirement for precise detection of keypoints in both images.
We adopt the semi-sparse paradigm and integrate it seamlessly with the coarse-to-fine architecture \cite{loftr}, 
allowing RCM to maintain efficiency while globally searching for sub-pixel matches in the target image.

To summarize, this paper presents the following contributions:
\begin{itemize}
	\item We seamlessly integrate the semi-sparse paradigm with the coarse-to-fine architecture, 
	establishing a strong foundation for our subsequent efforts to raise the theoretical upper bound of feature matching.
	\item We design a view switcher that dynamically switches the source and target images based on their scales.
	This view switcher offers a direct and effective method for increasing the number of matchable points in the source image.
	\item We propose a conflict-free coarse matching module, 
	which matches two sets of features in a many-to-one fashion,
	further breaking through the match quantity barrier in large scale variation scenes.
	\item The proposed RCM and RCM$_{\mathrm{Lite}}$ achieve an excellent performance-efficiency balance, 
	rendering the matchers suitable for a wider range of applications.
 \end{itemize}

\section{Related Works}
\label{sec:related}

\subsection{Local Feature Matching}
Local feature matching methods are categorized as sparse and dense methods.
Sparse methods involve the extraction of sparse features with a keypoint detector, 
followed by establishing correspondences through a matching process. 
Dense methods directly establish correspondences between dense features. 

\textbf{Sparse matching} contains well-known methods like SIFT \cite{sift}, SURF \cite{surf}, BRIEF \cite{brief} and ORB \cite{orb}, which are widely utilized.
The advent of deep learning introduces learning-based descriptors as a dominant approach. 
Learning-based detectors and descriptors \cite{d2net,superpoint,r2d2,dedode,aliked,silk} harness the power of convolutional neural networks (CNN) for improved robustness.
SuperGlue \cite{superglue} pioneered global relationship modeling among sparse features using Transformer \cite{transformer}, 
which enriches features via self- and cross-attention mechanisms. 

\textbf{Dense matching} methods \cite{loftr,matchformer,aspanformer} eschew the detector and instead establish dense correspondences between images, 
resulting in a significant improvement over sparse methods.
We attribute this advancement to the increase in ground-truth matches facilitated by coarse-to-fine matching, which effectively covers the entire target image.
This success in raising the feature matching ceiling inspires RCM to inherit the coarse-to-fine matching approach and pursue further enhancements in ground-truth matches.

\textbf{Semi-Sparse matching} methods, matching sparse and dense features asymmetrically,
are proposed by S2DHM \cite{s2dhm} and S2DNet\cite{s2dnet}.
These methods, while preserving sparse keypoints in the source image, 
use dense features at full image resolution in the target image, resulting in excessive memory usage.
In our approach, we integrate the semi-sparse matching paradigm with coarse-to-fine matching \cite{loftr} to maintain high efficiency.
Additionally, the proposed dynamic view switching and many-to-one matching empower the semi-sparse paradigm to achieve performance comparable to modern sparse and dense paradigms.

\subsection{Feature Matching with Scale Invariance}
Feature matching struggles in scenes with large scale variations,
primarily due to the scarcity of matchable points in small scale images and matching conflicts due to one-to-one matching.
Many methods are proposed to align scales before matching. 
OETR \cite{oetr} identifies overlapping regions between images with object detection. 
ScaleNet \cite{scalenet} and Scale-Net \cite{scale-net} directly predict the scale variation.
To address the scale challenge,
we divide scenes with large scale variations into two cases: 
source and target images with (1) small and large scales and (2) large and small scales.
In the first case, both sparse \cite{superglue,sgmnet} and dense \cite{loftr,matchformer} methods face a reduction in matchable points in the source image,
which is solved by the proposed dynamic view switcher.
In the second case, dense methods encounter matching conflicts in small scale target images, a challenge mitigated by the proposed many-to-one matching strategy.
Similar to many-to-one matching, AdaMatcher \cite{adamatcher} addresses the matching conflict in the second case through adaptive assignment. 
However, it is limited to one-to-one matching in the first case, where the scarcity of matchable points in the source image persists.
With the combination of dynamic view switcher and many-to-one matching, 
we can effectively obtain more matchable points in the source image and perform conflict-free matching in the target image in all scenes with large scale variations.

\section{Methodology}
\label{sec:methodology}

\begin{figure*}[t]
	\includegraphics[width=0.85\linewidth]{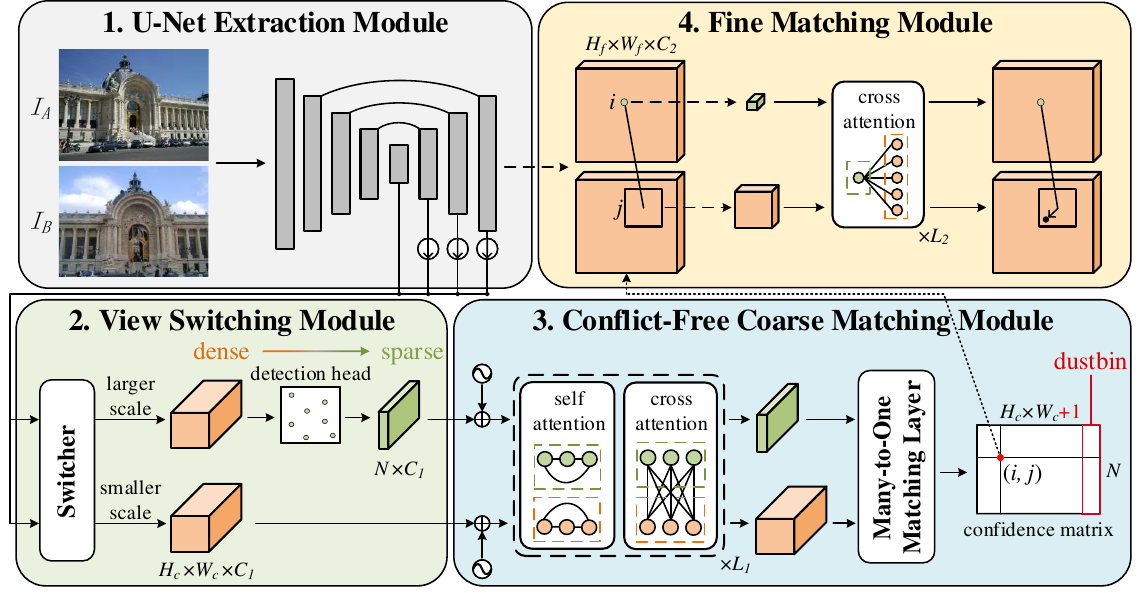}
	\centering
	 \caption{
		RCM is comprised of four primary modules: 
		\textbf{1.} The U-Net Extraction Module extracts multi-scale coarse and fine features.
		\textbf{2.} The View Switching Module dynamically switches the larger scale image to the sparse branch, 
		where dense coarse features are transformed into sparse features through the detection head.
		\textbf{3.} The Conflict-Free Coarse Matching Module involves attention layers processing two sets of coarse features, which are subsequently matched via the many-to-one matching layer.
		\textbf{4.} The Fine Matching Module further refines coarse matches based on fine features.
	 }
	 \label{figure3}
\end{figure*}

\subsection{Overview}
The overall architecture of RCM is illustrated in \cref{figure3}.
The U-Net extraction module initiates the process by employing CNN to extract coarse and fine features from both images. 
Subsequently, the coarse features of the two images undergo processing within the view switcher.
In this module, the coarse features of larger scale images are switched to the sparse branch and extracted by the detection head as sparse features.
And the coarse features of smaller scale images are retained as dense features.
The sparse and dense features are then matched in a many-to-one fashion through the conflict-free coarse matching module.
Finally the fine matching module crops out the fine features in the coarse matching positions and refines coarse matches through the correlation-based approach \cite{loftr}.

\subsection{Local Feature Extraction}
We extend the VGG-style encoder from SuperPoint \cite{superpoint} to a U-Net architecture to extract multi-scale features from images $I_{A}$ and $I_{B}$.
The decoder output is employed as fine features,
denoted as $F^{f}_{A} \in\mathbb{R}^{H_{f} \times W_{f} \times C_{2}}$ and $F^{f}_{B} \in\mathbb{R}^{H_{f} \times W_{f} \times C_{2}}$.
For brevity, we omit the $A$ and $B$ subscripts in the rest of the paper.
On the other hand, the coarse features, denoted as $F^{c} \in\mathbb{R}^{H_{c} \times W_{c} \times C_{1}}$, 
are formed by merging features from multiple resolutions, enhancing the robustness of the conflict-free coarse matching module.

\subsection{View Switching Module}

\begin{wrapfigure}[15]{r}{0.55\textwidth}
	\includegraphics[width=\linewidth]{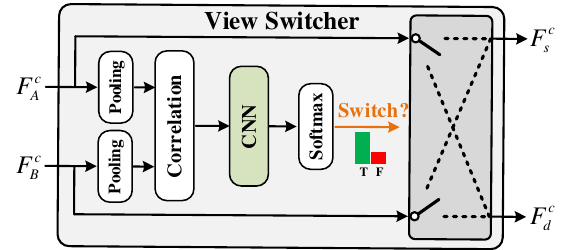}
	\centering
	\caption{The view switcher perceives the scale variation between image pairs
	to predict a binary classification result, indicating whether to switch the images.}
	 \label{figure4}
\end{wrapfigure}

As shown in \cref{fig:m2o_switcher_viz}(b), employing a small scale image as the source image leads to a scarcity of matchable points,
which consequently leads to fewer matches.
As shown in \cref{figure4}, we propose a binary classification network named view switcher to determine whether to switch the source and target images,
where a lightweight CNN is employed to process the correlation map of coarse features $F^{c}_{A}$ and $F^{c}_{B}$ to perceive scale variation.
The result of view switcher $VS$ is computed as
\begin{equation}
	\begin{split}
		& VS = \mathrm{Softmax}(\mathrm{CNN}(\mathrm{Corr}(\mathrm{Pool}(F^{c}_{A}), \mathrm{Pool}(F^{c}_{B})))),
		\label{switcher}
	\end{split}
\end{equation}
where $\mathrm{Pool}(\cdot)$ represents average pooling and $\mathrm{Corr}(\cdot,\cdot)$ signifies similarity calculation through the inner product.
The view switcher is trained to keep sparse branch processing larger scale images and dense branch processing smaller scale images.
In the sparse branch, we employ the detection head in \cite{superpoint} to detect keypoints and sample the feature map at the keypoint positions as sparse features $F^{c}_{s}$.
The U-Net encoder and detection head are inherited from SuperPoint \cite{superpoint} and are frozen, 
making the detection equivalent to SuperPoint.
Assuming that $N$ keypoints are detected, features in the sparse branch are reduced from $\mathbb{R}^{H_{c} \times W_{c} \times C_{1}}$ to $\mathbb{R}^{N \times C_{1}}$. 

\subsection{Conflict-Free Coarse Matching}
\subsubsection{Position Encoder}
Given that the attention mechanism lacks awareness of the positions of the features, we embed positional information into the coarse features $F^{c}$
through a position encoder $\mathrm{PE}$,
in which a lightweight $\mathrm{MLP}$ progressively expands the positional information $P\in\mathbb{R}^{N\times 3}$ into the coarse feature dimension $C_{1}$ and updates the coarse features. 
$P$ encompasses the $x$ and $y$ coordinates of keypoints, and their confidence scores $s$ generated by the detection head.

\subsubsection{Attention Layer}
The self- and cross-attention layers are employed for the global enhancement of sparse and dense features.
In contrast to dense methods that use linear attention \cite{linear_attn} for efficiency, the semi-sparse matching paradigm allows us to adopt vanilla attention \cite{transformer},
which is denoted as
\begin{equation}
	\begin{split}
		& M = \mathrm{Attention}(Q, K ,V)  = \mathrm{Softmax}(QK^{T})V.
		\label{equ2}
	\end{split}
\end{equation}
The attentional message is denoted as $M$.
In self-attention, $Q, K, V$ are three linear projections of a single feature set. 
In cross-attention, $Q$ is projected from one feature set, while $K$ and $V$ are projected from another feature set.

The feed-forward network $\mathrm{MLP_f}$ processes the attentional message $M$ to update the features as

\begin{equation}
	\begin{split}
		& F^{i+1} = F^{i} + \mathrm{MLP_f}(F^{i} || M),
		\label{equ3}
	\end{split}
\end{equation}
where $F^{i}$ is the feature in layer $i$ and $||$ is the concatenation operation.
We perform self- and cross-attention on sparse and dense features for $L_{1}$ iterations,
resulting in enhanced features $\tilde{F}_{s}^{c}$ and $\tilde{F}_{d}^{c}$ that capture global information.

\subsubsection{Many-to-One Coarse Matching with Dustbin}
With globally enhanced features, we perform matching between sparse and dense features in a many-to-one manner.
Initially, we utilize a set of learnable parameters as a dustbin, denoted as $bin\in\mathbb{R}^{1 \times C_{1}}$, and concatenate it with the dense features $\hat{F}_{d}^{c} = \tilde{F}_{d}^{c}||bin\in\mathbb{R}^{(H_{c}W_{c}+1) \times C_{1}}$. 
The dustbin is introduced to match with non-matchable sparse features, enabling our matcher to handle non-overlapping regions and occlusion.
We compute the score matrix with dustbin $S_{bin}$ between the sparse feature $\tilde{F}_{s}^{c}\in\mathbb{R}^{N \times C_{1}}$ and dense feature with dustbin $\hat{F}_{d}^{c}\in\mathbb{R}^{(H_{c}W_{c}+1) \times C_{1}}$ as
\begin{equation}
	\begin{split}
		& S_{bin} = \tilde{F}_{s}^{c}\hat{F}_{d}^{c}/\tau\in\mathbb{R}^{N \times (H_{c}W_{c}+1)},
		\label{equ4}
	\end{split}
\end{equation}
where $\tau$ is a learnable temperature parameter.

Current dense methods utilize the $\mathrm{Dual}$-$\mathrm{Softmax}$ operator to compute the coarse matching probability $P_{c}$, 
defined as $P_c = \mathrm{Softmax}_{row}(S) \cdot \mathrm{Softmax}_{col}(S)\in\mathbb{R}^{N \times H_{c}W_{c}}$.
The $\mathrm{Dual}$-$\mathrm{Softmax}$ operator keeps the summation of rows and columns of $P_{c}$ less than or equal to 1.
This limits each feature to match at most one feature, that is, one-to-one matching.
However multiple points may correspond to one point in scenes with large scale variations,
which leads to matching conflicts as shown in \cref{figure2}(b).
To address this issue, a single $\mathrm{Softmax}$ operation on the score with dustbin $S_{bin}$ is employed to release the constraint on target images while maintaining the constraint on source images as
\begin{equation}
	\begin{split}
		& P_c = \mathrm{Softmax}_{row}(S_{bin}),
		\label{equ5}
	\end{split}
\end{equation}
where each sparse feature can only match one dense feature or the dustbin, while one dense feature can match multiple sparse features.
The strategy of many-to-one matching with dustbin
can be comprehended as independently assigning each sparse feature to either its corresponding dense feature or to the dustbin. 
As shown in \cref{figure2}(c), 
following conflict-free coarse matching, the positions of sparse features, 
which correspond to the same dense feature, are independently adjusted during the fine matching stage.

Ultimately, features matched to the dustbin are discarded, and we proceed to select matches with matching probabilities surpassing the threshold $\theta_{c}$.

\subsection{Fine Matching}
For each coarse match, we sample a single feature at the source fine feature map 
and crop a feature window of size $w\times w$ at the target fine feature map.
Given the collection of $K$ coarse matches, we obtain fine features denoted as 
$\hat{F}_{s}^{f}\in\mathbb{R}^{K \times 1 \times C_{2}}$ and $\hat{F}_{d}^{f}\in\mathbb{R}^{K \times w^2 \times C_{2}}$
for the sparse and dense branches, respectively.

Subsequently, $\hat{F}_{s}^{f}$ and $\hat{F}_{d}^{f}$ are processed by the cross-attention module for $L_2$ iterations. 
Inspired by \cite{aspanformer}, in order to capture the local context of dense features, we substitute the self-attention module with the integration of $3 \times 3$ convolutions into the feed-forward network.
For fine matching, the features are updated by
\begin{equation}
	\begin{split}
		& F^{i+1} = F^{i} + \mathrm{Conv_3}(F^{i} || M).
		\label{equ6}
	\end{split}
\end{equation}

We compute the correlation map between the centroid of the source feature window and all the features in the target feature window, 
which represents the matching probability, and then get the fine matching position by computing the expectation of the probability distribution,
as introduced in \cite{loftr}.

\subsection{Supervision}
Our loss function encompasses four components: the coarse matching loss $\mathcal{L}_{c}^{m}$, 
the coarse dustbin loss $\mathcal{L}_{c}^{d}$, the fine matching loss $\mathcal{L}_{f}$, and the view switching loss $\mathcal{L}_{vs}$.
Ground-truth matches $M_{gt}=\{(i_{m}, j_{m})\}$ are generated from camera poses and depth maps, establishing many-to-one correspondence between sparse and dense features. 
And non-matchable sparse features $U_{gt}=\{i_{u}\}$, which are keypoints without any ground-truth match, are also obtained.

The coarse matching loss $\mathcal{L}_{c}^{m}$ is computed as the negative log-likelihood loss of the ground-truth matches $M_{gt}$ over the matching probability matrix $P_{c}$ of each attention layer,
\begin{equation}
	\mathcal{L}_{c}^{m} = - \frac{1}{L}\sum_{l}[\frac{1}{|M_{gt}|} \sum_{(i,j)\in M_{gt}} \log P_{c}^{l}(i, j)].
	\label{equ7} 
\end{equation}

The coarse dustbin loss $\mathcal{L}_{c}^{d}$ refers to the negative log-likelihood loss of the non-matchable sparse features $U_{gt}$ over the last column of the matching probability matrix $P_{c}$,
which signifies the matching probabilities with the dustbin,
\begin{equation}
	\mathcal{L}_{c}^{d} = - \frac{1}{|U_{gt}|} \sum_{i\in U_{gt}} \log P_{c}(i, H_{c}W_{c}+1).
	\label{equ8} 
\end{equation}

The fine matching loss $\mathcal{L}_{f}$ is computed by the L2-distance between each refined position and the ground-truth position, as introduced in \cite{loftr}.

To supervise the view switcher, we randomly select 500 ground-truth matches in each image pair and compute the distance between points in each image.
The image with the larger average distance is identified as the larger scale image. 
If the original source image is the larger scale image,
the ground-truth $VS_{gt}$ is false, and vice versa. 
The binary cross-entropy loss of the switcher result $VS$ is subsequently computed as
\begin{equation}
	\mathcal{L}_{vs} = - [VS_{gt}\log(VS)+(1-VS_{gt})\log(1-VS)].
	\label{equ9} 
\end{equation}
\section{Experiments}
\label{sec:experiments}
\subsection{Implementation Details}\label{4.1}
Both RCM and RCM$_\mathrm{Lite}$ consist of $L_{1}=5$ layers of coarse attention and $L_{2}=2$ layers of fine attention.
The dimensions for coarse and fine features are $C_{1}=256$ and $C_{2}=64$, respectively.
The key distinction between RCM and RCM$_\mathrm{Lite}$ lies in their coarse feature resolution and fine matching window size.
For RCM, the coarse feature resolution is $1/8$, and the fine matching window size is $w=5$.
For RCM$_\mathrm{Lite}$, the coarse feature resolution is $1/16$, and the fine matching window size is $w=9$. 
The learnable temperature $\tau$ is initialized to $0.1$, and the coarse matching threshold $\theta_{c}$ is set to $0.2$. 
The proposed method is implemented using Pytorch \cite{pytorch}.
During the training phase, images are resized and padded to the size of $832 \times 832$.
We detect $1024$ keypoints in the source image, and if there are insufficient keypoints, random keypoints are added to ensure efficient batching. 
RCM and RCM$_\mathrm{Lite}$ are trained on the MegaDepth dataset \cite{megadepth} for $30$ epochs with a batch size of $4$ by AdamW optimizer \cite{adamw} and are not fine-tuned for any other tasks.
The initial learning rate is set to $0.0002$, and the training employs the cosine decay learning rate scheduler with 1 epoch of linear warm-up.
More details can be found in the Supplementary Material.

\begin{table*}[t]
	\centering
	\caption{Evaluation on HPatches~\cite{hpatches} for homography estimation. 
	We report the percentage of correctly estimated homographies with mean corner errors below 1/3/5 pixels. We mark the best accuracy in \textbf{bold}.}
	\setlength{\tabcolsep}{6pt}
	\resizebox{0.99\linewidth}{!} {
	\begin{tabular}{c l | c c c  }
		\toprule
		\multirow{2}{*}{Category} &\multicolumn{1}{c|}{\multirow{2}{*}{Method}} &  Overall & Illumination & Viewpoint \\
		&  & \multicolumn{3}{c}{Accuracy ($\%,\epsilon< 1/3/5$ px)} \\
		\midrule
		\multirow{8}{*}{Sparse}  &SuperPoint~\cite{superpoint} + NN  & 0.46 / 0.78 / 0.85 & 0.57 / 0.92 / 0.97 & 0.35 / 0.65 / 0.74    \\
		&D2Net~\cite{d2net} + NN       & 0.38 / 0.72 / 0.81 & 0.65 / 0.95 / 0.98 & 0.13 / 0.51 / 0.65    \\
		&R2D2~\cite{r2d2} + NN         & 0.47 / 0.78 / 0.83 & 0.63 / 0.93 / 0.98 & 0.33 / 0.64 / 0.70     \\
		&ASLFeat~\cite{aslfeat} + NN & 0.48 / 0.81 / 0.88 & 0.63 / 0.94 / 0.98 & 0.34 / 0.69 / 0.78    \\
		&SuperPoint + SuperGlue~\cite{superglue} &  0.51 / 0.83 / 0.89 & 0.62 / 0.93 / 0.98 & 0.41 / 0.73 / 0.81  \\
		&SuperPoint + CAPS~\cite{caps} + NN& 0.49 / 0.79 / 0.86 & 0.62 / 0.93 / 0.98  & 0.36 / 0.65 / 0.75    \\                    
		&SIFT + CAPS~\cite{caps} + NN    & 0.36 / 0.76 / 0.85 & 0.48 / 0.89 / 0.95 & 0.26 / 0.65 / 0.76   \\
		&SuperPoint + LightGlue~\cite{lightglue} &  0.39 / 0.85 / 0.91 & 0.52 / 0.96 / 0.98 & 0.28 / 0.75 / 0.84  \\
		\midrule 
		\multirow{5}{*}{(Semi-)Dense}  &SparseNCNet~\cite{sparsencnet} & 0.36 / 0.66 / 0.76 & 0.62 / 0.92 / 0.97 & 0.13 / 0.41 / 0.57  \\
		&NCNet~\cite{ncnet}    & 0.48 / 0.61 / 0.71 & \textbf{0.98} / \textbf{0.98} / 0.98 & 0.02 / 0.28 / 0.46    \\
		&Patch2Pix~\cite{patch2pix}           & 0.50 / 0.79 / 0.87  & 0.71 / 0.95 / 0.98 & 0.30 / 0.64 / 0.76   \\
		&LoFTR~\cite{loftr}  & 0.55 / 0.81 / 0.86  & 0.74 / 0.95 / 0.98 & 0.38 / 0.69 / 0.76   \\
		&MatchFormer~\cite{matchformer}  & 0.55 / 0.81 / 0.87  & 0.75 / 0.95 / 0.98 & 0.37 / 0.68 / 0.78   \\
		\midrule
		\multirow{2}{*}{Semi-Sparse}  	& RCM$_{\mathrm{Lite}}$  & 0.50 / 0.86 / 0.91  & 0.55 / 0.95 / 0.98 & 0.45 / 0.79 / 0.85   \\
		& RCM  & \textbf{0.56} / \textbf{0.88} / \textbf{0.93}  & 0.65 / 0.97 / \textbf{0.99} & \textbf{0.47} / \textbf{0.80 / 0.88}   \\
		\bottomrule	
	\end{tabular}
	}
	\label{tab:exp_homography}
\end{table*}

\subsection{Homography Estimation}\label{4.3}
\textbf{Dataset.}
The HPatches \cite{hpatches} dataset is widely recognized for evaluating local features and matchers. 
In line with the evaluation protocol in \cite{d2net}, we assess the performance of our method across 108 HPatches sequences, 
which comprise 52 sequences with illumination variations and 56 sequences with viewpoint changes.

\noindent\textbf{Metric.}
Following the metrics employed in \cite{superpoint,superglue,patch2pix}, 
we report the percentage of correctly estimated homographies with mean corner errors below 1/3/5 pixels.
Vanilla RANSAC \cite{ransac} is applied to estimate the homography for all methods.

\noindent\textbf{Results.}
As shown in \cref{tab:exp_homography}, RCM achieves the best overall performance at all thresholds.
Comparing the dense methods, RCM shows a significant advantage in accuracy in large viewpoint change scenarios, 
leading LoFTR and MatchFormer by (+9\%,+11\%,+12\%) and (+10\%,+12\%,+10\%), respectively.
We attribute this performance improvement to the increased matchable points in the source image facilitated by the view switcher, and the conflict-free matching achieved in the target image through many-to-one matching.
Compared to the sparse method SuperGlue, RCM achieves an overall performance lead of (+5\%,+5\%,+4\%), 
benefiting from the dense search due to the semi-sparse paradigm and the sub-pixel accuracy due to the coarse-to-fine architecture.

\begin{table}[ht]
	\centering
	\begin{minipage}[c]{0.49\textwidth}
		\centering
		\captionof{table}{Two-view pose estimation results on MegaDepth \cite{megadepth} in outdoor scenes.}
			\label{tab:MegaDepth_result}
			\setlength{\tabcolsep}{6pt}
			\resizebox{0.99\linewidth}{!}{
				\begin{tabular}{c l | c c c}
					\toprule
					\multirow{2}{*}{Category} &\multicolumn{1}{c|}{\multirow{2}{*}{Method}} &\multicolumn{3}{c}{Pose estimation AUC} \\
							& &{@$5^\circ$}  &{@$10^\circ$}  &{@$20^\circ$} \\
					\midrule
					\multirow{6}{*}{Sparse} &SuperGlue~\cite{superglue} &42.2 &59.0 &73.6 \\
					
													&SGMNet~\cite{sgmnet} &40.5 &59.0 &73.6 \\
		
													&ClusterGNN~\cite{clustergnn} &44.2 &58.5 &70.3 \\
		
													&LightGlue~\cite{lightglue} &46.4 &64.3 &78.2 \\

													&DeDoDe-B~\cite{dedode} &49.1 &65.0 &77.1 \\

													&DeDoDe-G~\cite{dedode} &51.2 &68.4 &81.2 \\
					\midrule
					\multirow{4}{*}{Dense} &DRC-Net~\cite{dual} &27.0 &42.9 &58.3 \\
													
													&PDC-Net+(H)~\cite{pdcnet}  &43.1  &61.9  &76.1 \\
													
													&LoFTR~\cite{loftr} &52.8 &69.2 &81.2 \\     
													
													&RoMa~\cite{roma} &63.0 &76.9 &86.4 \\
					\midrule
					\multirow{2}{*}{Semi-Sparse} & RCM$_{\mathrm{Lite}}$ & 49.7 & 66.5 & 79.3 \\
					
					& RCM       &53.2 &69.4 &81.5  \\
					\bottomrule
				\end{tabular}
			}
	\end{minipage}
	\hfill
	\begin{minipage}[c]{0.49\textwidth}
		\centering
		\captionof{table}{Two-view pose estimation results on ScanNet \cite{scannet} in indoor scenes.}
		\label{tab:ScanNet_result}
		\setlength{\tabcolsep}{6pt}
		\resizebox{0.97\linewidth}{!}{
			\begin{tabular}{c l | c c c}
				\toprule
				\multirow{2}{*}{Category} &\multicolumn{1}{c|}{\multirow{2}{*}{Method}} &\multicolumn{3}{c}{Pose estimation AUC} \\
						& &{@$5^\circ$}  &{@$10^\circ$}  &{@$20^\circ$} \\
				\midrule
				\multirow{4}{*}{Sparse} 
												&OANet~\cite{oanet} &11.8 &26.9 &43.9 \\
												
												&SuperGlue~\cite{superglue} &16.2 &33.8 &51.8 \\
	
												&SGMNet~\cite{sgmnet} &15.4 &32.1 &48.3 \\  
	
												&LightGlue~\cite{lightglue}* &13.9 &29.6 &45.7\\
				\midrule
				\multirow{3}{*}{Dense}  &DRC-Net~\cite{dual}* &7.7 &17.9 &30.5 \\                         
				
												&MatchFormer~\cite{matchformer}* &15.8  &32.0   &48.0 \\
	
												&LoFTR-OT~\cite{loftr}* &16.9 &33.6 &50.6 \\
				\midrule
				\multirow{2}{*}{Semi-Sparse} & RCM$_{\mathrm{Lite}}$* & 15.2 & 31.5 & 48.8 \\
				
				& RCM*       &17.3 &34.6 &52.1  \\
				\bottomrule
			\end{tabular}
		}
	\end{minipage}
\end{table}

\subsection{Relative Pose Estimation}\label{4.4}
\textbf{Dataset.}
We evaluate matchers on the MegaDepth \cite{megadepth} and the ScanNet \cite{scannet} dataset for outdoor and indoor pose estimation, respectively.
For the MegaDepth dataset, all test images are resized such that their longer dimensions are $1216$.
For the ScanNet dataset, all test images are resized to $640\times480$ pixels.

\noindent\textbf{Metric.}
Based on the matching results, we apply vanilla RANSAC to solve the fundamental matrices and relative poses of the image pairs.
The AUC of the pose error at thresholds $(5^\circ, 10^\circ, 20^\circ)$ are reported, where pose error is the maximum angular error in rotation and translation.

\begin{figure*}[t]
	\includegraphics[width=0.99\linewidth]{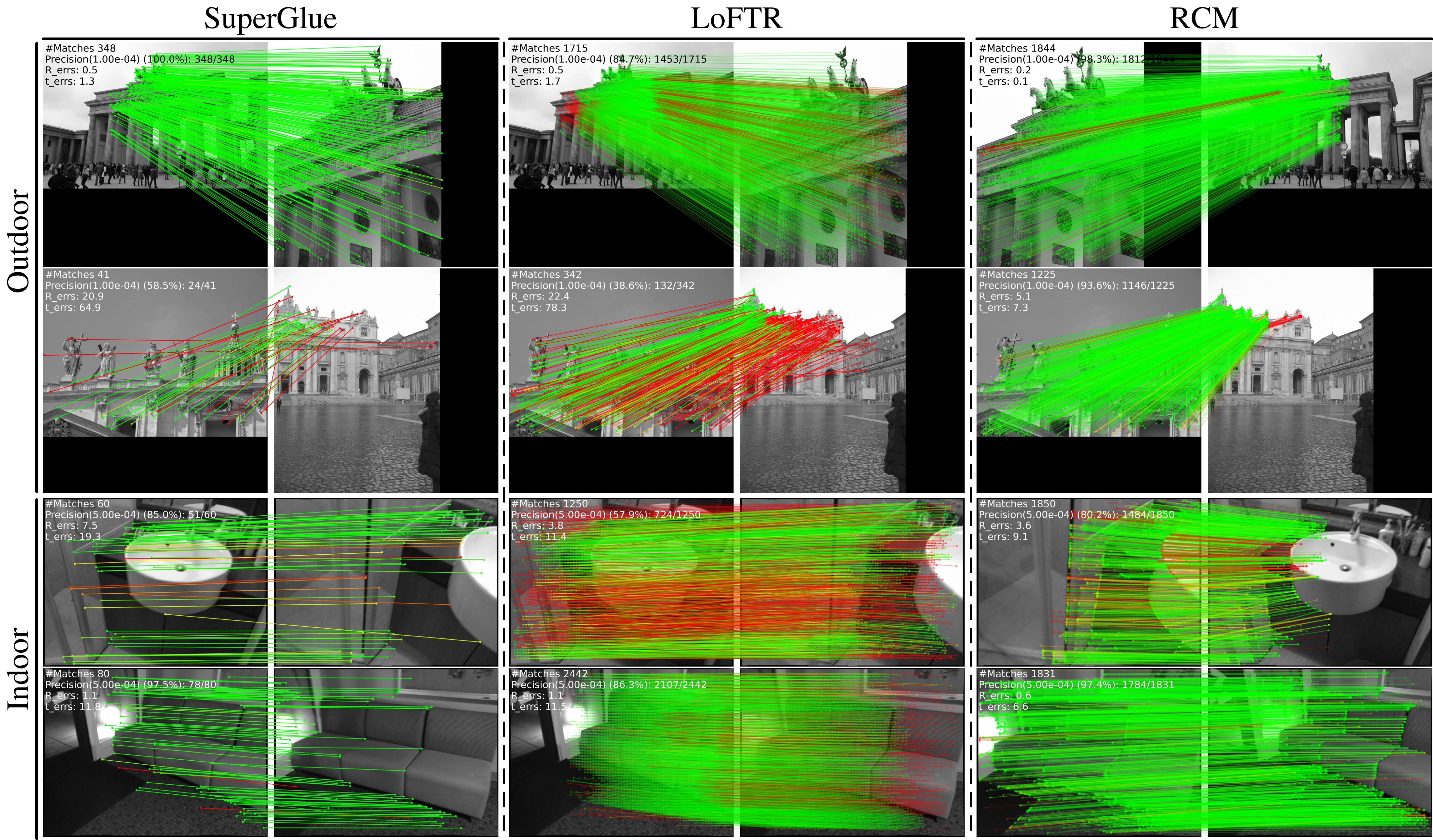}
	\centering
	 \caption{Qualitative comparison of SuperGlue \cite{superglue}, LoFTR \cite{loftr} and proposed RCM on MegaDepth and ScanNet.}
	 \label{fig:viz}
\end{figure*}

\noindent\textbf{Results.}
As shown in \cref{tab:MegaDepth_result}, RCM achieves the second best performance on the MegaDepth dataset.
RCM outperforms the state-of-the-art method LoFTR, demonstrating the benefits of the view switcher and many-to-one matching.
The lightweight RCM$_{\mathrm{Lite}}$ achieves a performance boost $(+7.5\%, +7.5\%, +5.7\%)$ compared to SuperGlue.
\cref{fig:viz} illustrates qualitative comparisons of the SuperGlue, LoFTR and RCM. 
More visualizations and analyses can be found in the Supplementary Material.

As shown in \cref{tab:ScanNet_result}, RCM achieves a $(0.4\%, 1\%, 1.5\%)$ performance improvement over LoFTR in the ScanNet dataset.
It's worth noting that the superscript $*$ indicates zero-shot performance, and the model is trained on MegaDepth. 
Nevertheless, our methods achieve performance comparable to state-of-the-art sparse methods trained on ScanNet.
Dense and semi-sparse matching methods find matches in the target image exhaustively, mitigating the challenge of non-repeatability in indoor scenes, a key limitation of sparse methods.

\subsection{Visual Localization}
\textbf{Dataset.}
We assess the performance of our method using the Aachen Day-Night v1.1 benchmark \cite{aachen}, a well-established dataset for visual localization.
This dataset comprises 824 day-time images and 191 night-time images, 
chosen as queries for outdoor visual localization. 

\noindent\textbf{Metric.}
We utilize the open-source localization pipeline HLoc \cite{hloc} for conducting localization with RCM.
We report the percentage of successfully localized images within three error thresholds $(0.25m, 2^\circ)$ / $(0.5m, 5^\circ)$ / $(1m, 10^\circ)$.

\noindent\textbf{Results.}
As shown in \cref{tab:Aachen_result},
RCM demonstrates comparable or superior performance compared to the dense methods LoFTR and ASpanFormer in outdoor visual localization tasks.
Our approach exhibits strong generalization across various downstream tasks without the need for fine-tuning.
It is noteworthy that the semi-sparse method RCM is significantly more efficient than the dense methods, achieving a runtime speedup of over twofold.

\begin{table}[t]
	\centering
	\caption{Visual localization on the Aachen Day-Night benchmark v1.1~\cite{aachen}.}
	\label{tab:Aachen_result}
	\setlength{\tabcolsep}{8pt}
	\resizebox{0.7\linewidth}{!}{
		\begin{tabular}{ l c c}
			\hline
            \multicolumn{1}{c}{\multirow{2}{*}{Method}} & Day & Night \\
			\cline{2-3}
                     & \multicolumn{2}{c}{$\left(0.25m, 2^\circ\right)$ / $\left(0.5m, 5^\circ\right)$ / $\left(1m, 10^\circ\right)$} \\
			\hline
			\multicolumn{3}{l}{\textbf{Localization with matching pairs generated by HLoc}} \\
			\hline
            LoFTR~\cite{loftr} & 88.7 / 95.6 / 99.0 & \textbf{78.5} / 90.6 / 99.0 \\
            ASpanFormer~\cite{aspanformer} & 89.4 / 95.6 / 99.0 & 77.5 / \textbf{91.6} / 99.0 \\
			DeDoDe-B~\cite{dedode} & 87.4 / 94.7 / 98.5 & 70.7 / 88.0 / 97.9    \\
			DeDoDe-G~\cite{dedode} & 88.5 / 95.4 / 99.0 & 74.9 / 89.5 / \textbf{99.5}    \\
			SP~\cite{superpoint}+LightGlue~\cite{lightglue} & 89.6 / 95.8 / \textbf{99.2} & 72.8 / 88.0 / 99.0    \\
			 RCM$_\mathrm{Lite}$ & 87.7 / 94.7 / 98.4 & 72.3 / 91.1 / 98.4 \\
			 RCM & \textbf{89.7} / \textbf{96.0} / 98.7 & 72.8 / \textbf{91.6} / 99.0 \\
			\hline
		\end{tabular}
	}
\end{table}

\begin{figure}[t]
	\includegraphics[width=0.99\linewidth]{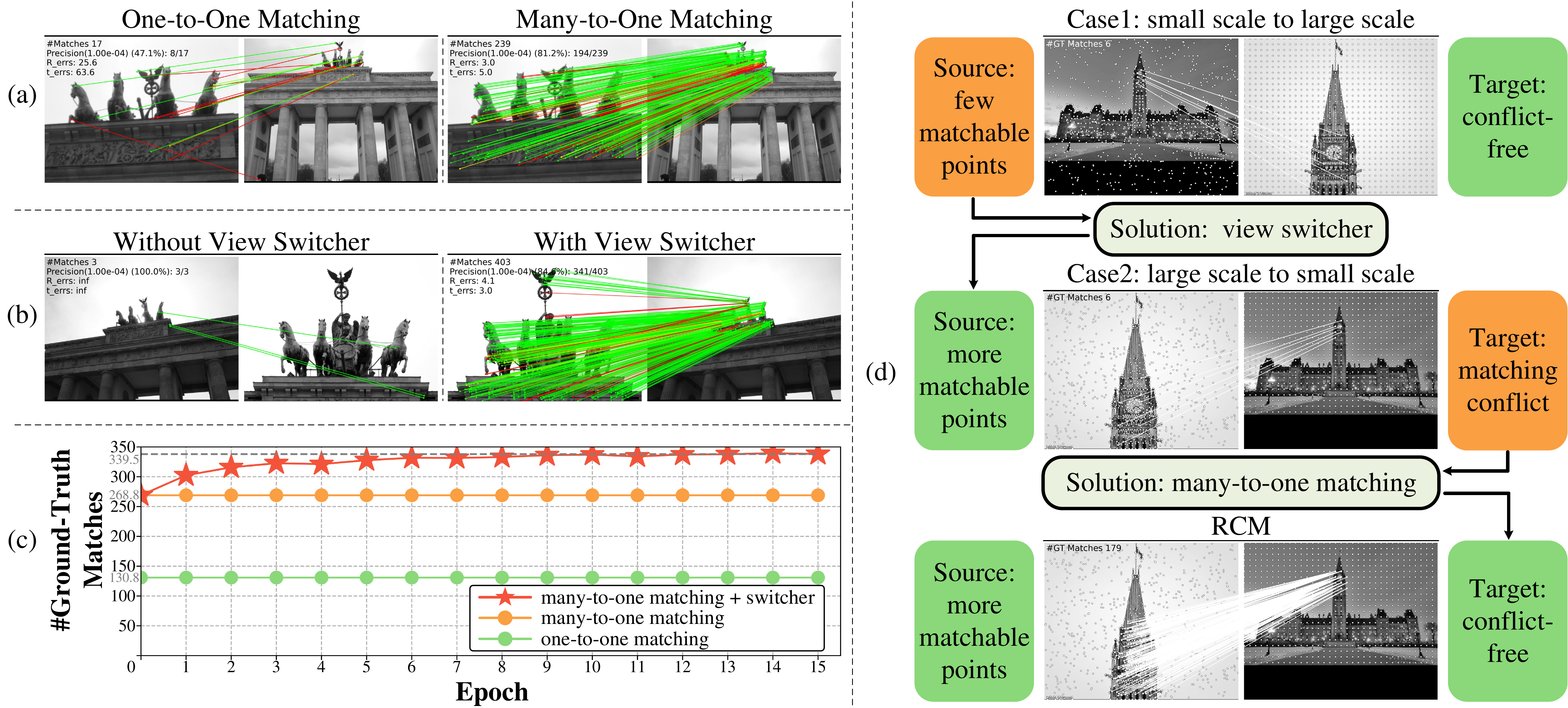}
	\centering
	 \caption{Ablation study on view switcher and many-to-one matching. (a) (b) shows the actual matching results and (c) (d) shows the ground-truth coarse matches.
	 The harmonious combination of view switcher and many-to-one matching allows more matchable points in the source image and conflict-free matching in the target image.
	 }
	 \label{fig:m2o_switcher_viz}
\end{figure}

\subsection{Ablation Study}\label{4.5}
In this subsection, we conduct qualitative and quantitative ablation experiments on MegaDepth.
All models in \cref{tab:ablation_result} are trained for 15 epochs at 544 resolution and tested at 1216 resolution.
For the baseline model (1), we apply the original SuperPoint \cite{superpoint} to generate keypoints and descriptors for both images.
O2O and M2O in \cref{tab:ablation_result} are abbreviations for one-to-one and many-to-one, respectively.

\noindent\textbf{Ground-Truth Match Analysis.}
\cref{fig:m2o_switcher_viz}(d) illustrates the core idea of RCM, 
which is to increase the theoretical upper bound on the matcher, \ie, ground-truth matches, through view switcher and many-to-one matching.
\cref{fig:m2o_switcher_viz}(c) illustrates a quantitative comparison of ground-truth matches,
showing an increase as the view switching accuracy improves during training.
Eventually, RCM and RCM$_{\mathrm{Lite}}$ achieve view switching accuracies of $90.9\%$ and $87.8\%$, respectively.
Through the integration of many-to-one matching and the view switcher, the ground-truth matches surge to $260\%$ compared to one-to-one matching,
which is a compelling demonstration of raising the ceiling of feature matching.

\begin{table}[ht]
	\centering
	\begin{minipage}[c]{0.49\textwidth}
		\centering
		\captionof{table}{Ablation study of each proposal on MegaDepth.}
		\label{tab:ablation_result}
			\resizebox{0.99\linewidth}{!}{
				\begin{tabular}{l | c c}
				\toprule
				Method  &\#Matches &Prec.  \\
				\midrule
				(1) Sparse sampling &451.7 &98.0  \\
				(2) Dense sampling &932.1 &93.5  \\
				(3) Semi-Sparse sampling &492.3 &93.7 \\
				\midrule
				(4) + U-Net Extraction      & 507.8	& 94.2   \\
				(5) O2O $\rightarrow$ M2O     & 770.7$_{(+51.8\%)}$    & 97.2    \\
				(6) + Switcher (RCM$_{\mathrm{Lite}}$)       & 951.4$_{(+23.4\%)}$     & 97.8     \\ 
				(7) $1/16$ $\rightarrow$ $1/8$ (RCM)       & \textbf{961.4}     & \textbf{98.1}   \\ 	
				\bottomrule
				\end{tabular}
				}
	\end{minipage}
	\hfill
	\begin{minipage}[c]{0.49\textwidth}
		\centering
		\captionof{table}{Evaluation on extreme-scale MegaDepth.}
		\label{tab:MegaDepth_scale}
		\resizebox{0.8\linewidth}{!}{
			\begin{tabular}{l | c c}
				\toprule
				Method  &\#Matches &Prec.  \\
				\midrule
				SP~\cite{superpoint}+SuperGlue~\cite{superglue} &261 &75.5 \\
				SP~\cite{superpoint}+LightGlue~\cite{lightglue} &313 &76.0 \\
				
				\midrule
				LoFTR~\cite{loftr} &424 &53.5 \\
				MatchFormer~\cite{matchformer} &464 &77.0 \\
				ASpanFormer~\cite{aspanformer} &425 &77.1 \\

				\midrule
				RCM       &\textbf{887}  &\textbf{78.9}   \\
				wo switcher      &653  &75.4   \\
				flipped switcher  &76  &54.2   \\
				\bottomrule
			\end{tabular}
		}
	\end{minipage}
\end{table}

\noindent\textbf{Sampling Strategy.}
As shown in \cref{tab:ablation_result}, 
the baseline models (1), (2), and (3) differ solely in sampling strategies for coarse features.
Precisely, (1) detects keypoints in both images, (2) employs dense feature as the coarse feature, 
and (3) detects keypoints in source image and employs dense feature in target image.

\noindent\textbf{Many-to-One Matching.}
As shown in \cref{fig:m2o_switcher_viz}(a), in contrast to the one-to-one matching strategy, the many-to-one matching strategy pursues conflict-free matching, resulting in more detailed matches.
As shown in \cref{tab:ablation_result}(5), substituting one-to-one matching with many-to-one matching improves the number of matches by $51.8\%$ and precision by $3\%$. 

\noindent\textbf{View Switcher.}
As shown in \cref{fig:m2o_switcher_viz}(b), without the view switcher, the matcher can only produce a small number of matches in small scale images 
because only a limited number of keypoints are detected in the overlapping region.
The view switcher enables the detection of more matchable keypoints in the source image. 
As shown in \cref{tab:ablation_result}(6), the average number of matches increases by $23.4\%$ with the view switcher.
More analyses about view switcher and many-to-one matching can be found in the Supplementary Material.

\noindent\textbf{RCM and RCM$_{\mathrm{\mathbf{Lite}}}$.}
The distinctions between RCM and RCM$_{\mathrm{Lite}}$ lie in the coarse feature resolution and fine matching window. 
RCM employs coarse features at $1/8$ resolution, consistent with the dense attention-based methods.
We apply $1/16$ resolution coarse features in RCM$_{\mathrm{Lite}}$ to trade performance degradation for faster runtime, facilitating applications in real-time scenarios.
Because of the additional layers at $1/16$ resolution in the feature extraction network, RCM$_{\mathrm{Lite}}$ consists of more parameters than RCM.
Also to cover larger coarse feature patches, we design a larger fine matching window for RCM$_{\mathrm{Lite}}$.

\noindent\textbf{Robustness to Scale Variation.}
We evaluate matchers on the MegaDepth subset defined in OETR \cite{oetr}, where the overlap scale ratio is in $[4,+\infty]$.
As shown in \cref{tab:MegaDepth_scale}, RCM achieves the best performance with $2.8\times$ and $2.1\times$ more matches than LightGlue and ASpanFormer respectively.
The performance of RCM is severely degraded without the view switcher,
demonstrating the significance of the view switcher in large scale variation scenarios.
Flipping the switching result, \ie, actively taking small scale image as source image, leads to the worst results.

\begin{wraptable}[10]{r}{0.6\textwidth}
	\caption{Efficiency Analysis on RTX3090.
	The \colorbox{red2}{1st}, \colorbox{orange}{2nd}, and \colorbox{yellow}{3rd}-best methods are highlighted.}
	\centering
	  \resizebox{\linewidth}{!}{
	  \begin{tabular}{l l | c c c | c}
		\toprule
		Category &Method &AUC@$10^\circ$  &\#Param. &Time (ms) &Avg. Rank\\
		\midrule
		\multirow{2}{*}{Sparse} & DeDoDe \cite{dedode} &65.0 &28.1M &322  &4.7 \\
		& LightGlue \cite{lightglue} &64.3 &11.9M &\cellcolor{red2}72 &\cellcolor{yellow}3.7\\
		\midrule
		\multirow{2}{*}{Dense} &LoFTR \cite{loftr} &\cellcolor{yellow}69.2 &\cellcolor{yellow}11.6M &454 &\cellcolor{yellow}3.7\\
		&RoMa \cite{roma} &\cellcolor{red2}76.9 &111M &913 &4.3\\
		\midrule  
		\multirow{2}{*}{Semi-Sparse} 
		&RCM$_{\mathrm{Lite}}$ &66.5 &\cellcolor{orange}11.4M  &\cellcolor{orange}134 &\cellcolor{orange}2.7\\
		&RCM &\cellcolor{orange}69.4  &\cellcolor{red2}9.8M &\cellcolor{yellow}208  &\cellcolor{red2}2\\
		\bottomrule
	  \end{tabular}}
	\label{tab:mega}
\end{wraptable}

\noindent\textbf{Efficiency.}
The parameters, runtimes, and performances of state-of-the-art matchers are presented in \cref{tab:mega}. 
Notably, RCM and RCM$_{\mathrm{Lite}}$ achieve the optimal balance between performance and efficiency.
Although the performance of RCM lags behind the most precise matcher RoMa, 
RCM achieves respectable performance with only $8.8\%$ of the parameters and $23\%$ of the runtime, making it more suitable for real-time applications.

\section{Limitation and Discussion}
\label{sec:diss}
The view switcher assumes that one image has a relatively large scale, 
while both images being small in scale remains a tricky case in feature matching.
A potential approach involves dynamically scaling the image pairs according to the overlapping regions, 
ensuring that the overlapping regions in two images remain simultaneously large and comparable in scale. 

Since the matching points in the target image are regression-based points rather than pre-detected points, 
RCM suffers from the lack of discrete tracks in the SfM task and requires additional adaptation strategy.

The semi-sparse paradigm mitigates the keypoint-repeatability reliance, addressing only a portion of the broader keypoint dependence.
Challenges in detection persist, such as low quantity and poor distribution, influencing the performance of RCM in texture-less scenes.
We believe that the integration of the semi-sparse paradigm with advanced detection methods
holds promising potential as a path forward for effective and efficient feature matching.

\section{Conclusion}
\label{sec:conclusion}
This paper introduces a novel perspective to enhance matchers by raising the theoretical upper bound.
We propose the dynamic view switching mechanism for the source image and the conflict-free coarse matching layer for the target image, respectively.
Through the integration of the view switcher and many-to-one matching layer, we enhance the number of matchable points in the source image while achieving conflict-free matching in the target image.
By essentially augmenting the number of ground-truth matches between images, RCM raises the theoretical upper bound on the actual matching results.
The results from qualitative and quantitative experiments demonstrate a significant enhancement in the practical performance of the matcher with a higher ceiling, 
particularly evident in challenging scenarios characterized by large scale variations.
With its outstanding performance and efficiency, we believe that RCM holds considerable potential for research and application.

\bibliographystyle{splncs04}
\bibliography{main}

\appendix
\newcommand{\AppendixPrefix}{S}
\renewcommand{\thefigure}{\AppendixPrefix\arabic{figure}}
\setcounter{figure}{0}
\renewcommand{\thetable}{\AppendixPrefix\arabic{table}}
\setcounter{table}{0}
\renewcommand{\theequation}{\AppendixPrefix\arabic{equation}}
\setcounter{equation}{0}
\setcounter{page}{1}

\title{Raising the Ceiling: Conflict-Free Local Feature Matching with Dynamic View Switching} 

\titlerunning{RCM: Conflict-Free Local Feature Matching with Dynamic View Switching}

\author{Supplementary Material}

\authorrunning{X.~Lu et al.}

\institute{}

\maketitle

The supplementary material is summarized as follows: 
Section \ref{sec:experiments_supp} presents additional qualitative and quantitative experiments.
Section \ref{sec:details_supp} offers additional details and insights into the design of networks.

\begin{center}
	\includegraphics[width=0.99\linewidth]{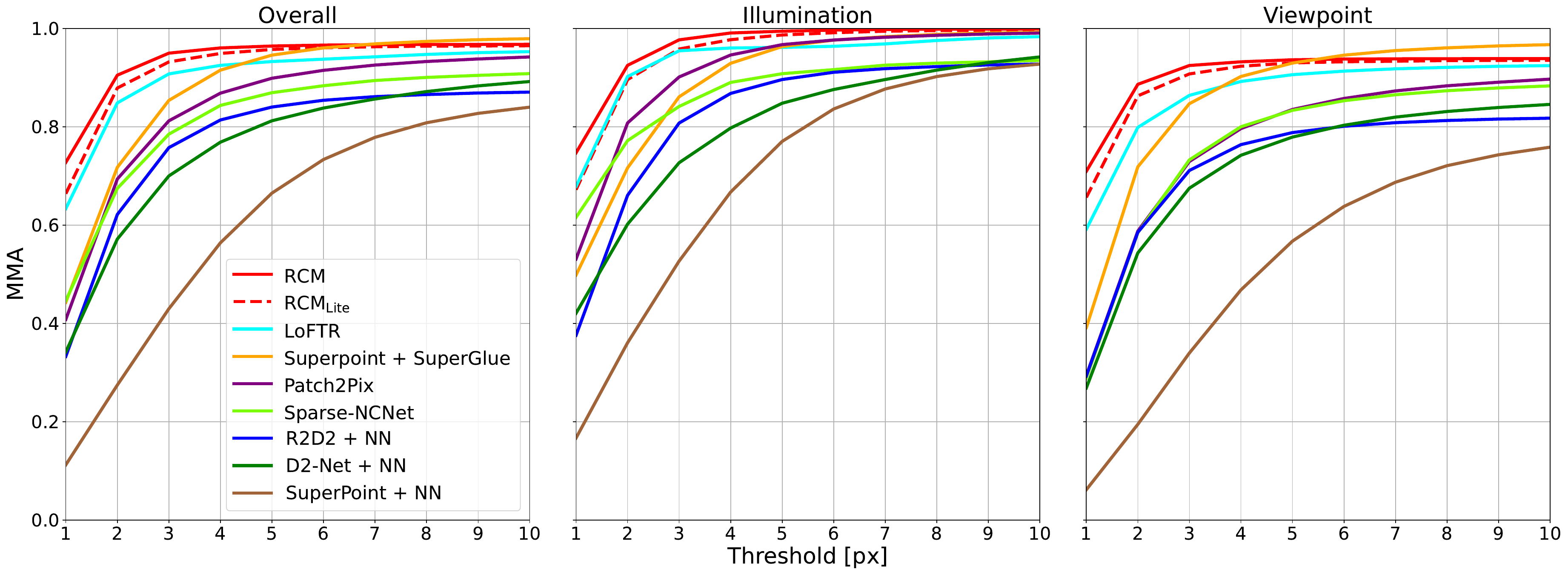}
	\captionof{figure}{\textbf{Image matching on HPatches~\cite{hpatches}.} MMA curves are plotted by changing the reprojection error threshold.}
	\label{fig:exp_matching}
\end{center}
	
\section{Additional Experiments}
\label{sec:experiments_supp}
\subsection{Image Matching}
\textbf{Dataset.}
Following the evaluation protocol introduced in D2-Net \cite{d2net}, 
we evaluate the performance of our method over 108 HPatches \cite{hpatches} sequences, 
which include 52 instances with illumination variations and 56 instances with viewpoint changes.

\noindent\textbf{Metric.}
We compute the reprojection error of each match from the homographies provided by the HPatches dataset. 
The matching threshold is varied from 1 to 10 to visualize the mean matching accuracy (MMA), which is the average percentage of correct matches for each image.

\noindent\textbf{Results.}
As shown in \cref{fig:exp_matching}, our method RCM achieves the best accuracy at thresholds less than or equal to 6,
and the RCM$_\mathrm{Lite}$ outperforms the dense method LoFTR at all thresholds.
Dense and semi-sparse methods demonstrate significantly superior accuracy at lower matching thresholds. 
This advantage stems from their ability to produce precise matches at the sub-pixel level within the target image, independent of the imprecision of the keypoints.
Compared to the dense method LoFTR, the semi-sparse matching paradigm with many-to-one matching and switcher 
can further improve the matching accuracy as it yields more precise matching points in the source image through detection.

\subsection{More Qualitative Results}

\noindent\textbf{Qualitative Ablation on Coarse Ground-Truth Matches.}
In \cref{fig:gt_match_conflict} and \cref{fig:gt_match_matchable},
we illustrate how many-to-one matching and the switcher benefit both dense and semi-sparse matching paradigms 
by resolving matching conflicts in the target image and increasing matchable points in the source image.
The substantial increase in ground-truth matches enhances the theoretical upper bound on the actual matching results produced by the matcher.

\noindent\textbf{Qualitative Ablation on Actual Matching Results.}
Additional comparisons between one-to-one and many-to-one matching strategies are presented in \cref{fig:m2o_switcher_viz_supp}. 
The many-to-one matching strategy substantially increases the number of matches and enhances accuracy by resolving conflicts in the coarse matching phase.

In \cref{fig:m2o_switcher_viz_supp}, further ablation comparisons for the switcher are provided. 
These comparisons consistently demonstrate that the switcher significantly increases the number of matches, 
consequently enhancing the performance of downstream tasks such as pose estimation.

\begin{figure}[t]
	\includegraphics[width=0.65\linewidth]{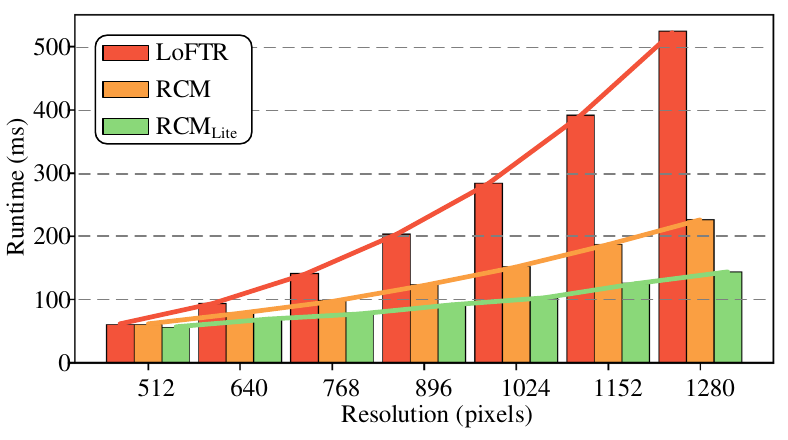}
	\centering
	 \caption{Efficiency Analysis.}
	 \label{fig:efficiency}
\end{figure}

\textbf{Qualitative Comparisons of Sparse, Dense and Semi-Sparse Methods.}
In \cref{fig:viz_supp2}, we present additional qualitative comparisons of three matching paradigms in both outdoor and indoor scenes.
In outdoor scenes, the superiority of our approach over sparse \cite{superglue} and dense \cite{loftr} methods in terms of accuracy and match quantity is evident from the first row.
This advantage stems from the semi-sparse paradigm, which extracts precise keypoints from the source image and conducts a global search within the target image. 
The last two rows highlight that our proposed switcher, responsible for switching larger scale images to the source image, significantly enhances the number of matches. 
This improvement is attributed to our improved ability to detect more matchable keypoints within the overlapping region. 

In indoor scenes, the proposed semi-sparse matching method, RCM, consistently produces superior results. 
In contrast to the sparse method, RCM achieves a significantly higher number of matches by mitigating reliance on keypoint repeatability, 
leading to improved pose estimation performance. 
Compared to the dense method, RCM excels in detecting keypoints at more discriminative positions, resulting in significantly higher matching precision. 

Additional indoor and outdoor qualitative comparisons are presented in \cref{fig:viz_wheel_in_supp} and \cref{fig:viz_wheel_out_supp}, where matched points are color-coded for clarity.

\noindent\textbf{Efficiency.}
As shown in \cref{fig:efficiency},
the runtime of the dense method LoFTR increases quadratically with image size.
Semi-Sparse paradigm mitigate the quadratic increase in the runtime with the detection in the source image.

\noindent\textbf{Visualizations of the Dustbin and Attention Weights.}
\cref{fig:viz_dustbin_attnn_supp}(a) illustrates the role of the dustbin, 
designed to discard non-matchable points in non-overlapping regions, 
enabling RCM to effectively handle occlusions and viewpoint changes.
Additionally, visualizations of self-attention weights and cross-attention weights are provided in \cref{fig:viz_dustbin_attnn_supp}(b) and (c), respectively.

\noindent\textbf{Failure Cases.}
We present the failure cases of RCM in \cref{fig:fail_supp}. 
These instances occur in outdoor scenes with severe scale changes and misclassification of the switcher. 
Failures also occur in indoor scenes featuring extensive texture-less areas and substantial viewpoint changes,
where the detector struggles to produce discriminating keypoints.

\noindent\textbf{3D Reconstruction Results.}
The HLoc pipeline \cite{hloc} is employed for 3D reconstruction based on the matching results of RCM, 
followed by dense reconstruction using COLMAP\cite{sfm}.
The sparse and dense models of three landmarks are illustrated in \cref{fig:viz_reconstruction_supp}.

\section{More Details}
\label{sec:details_supp}

\textbf{U-Net Feature Extraction.}
The encoder of the U-Net network inherits the SuperPoint \cite{superpoint} encoder, 
producing feature maps at resolutions of $1/2$, $1/4$, and $1/8$. 
In RCM, we design a similar VGG-like structure for the decoder, 
progressively integrating information from the encoder to generate fine features at $1/2$ resolution.
RCM$_\mathrm{Lite}$ takes an additional step by incorporating $1/16$ resolution encoding and decoding layers, 
accounting for its increased parameter count compared to RCM. 
RCM combines $1/2$, $1/4$, and $1/8$ resolution decoder features linearly to form coarse features, 
while RCM$_\mathrm{Lite}$ includes $1/2$, $1/4$, $1/8$, and $1/16$ resolution decoder features to generate coarse features. 
The feature dimensions of the $1/2$, $1/4$, $1/8$, and $1/16$ resolution maps are $64$, $128$, $256$, and $256$, respectively.

\noindent\textbf{Detection.}
In outdoor scenes, we detect keypoints with NMS radius of $4$ pixels and keypoint threshold of $0.005$, which are the default settings of SuperPoint \cite{superpoint}.
In indoor scenes, we adjust the parameters to an NMS radius of $1$ pixel and a keypoint threshold of $0.001$ to enhance keypoint detection in low-texture areas.

\begin{figure}[t]
	\includegraphics[width=0.99\linewidth]{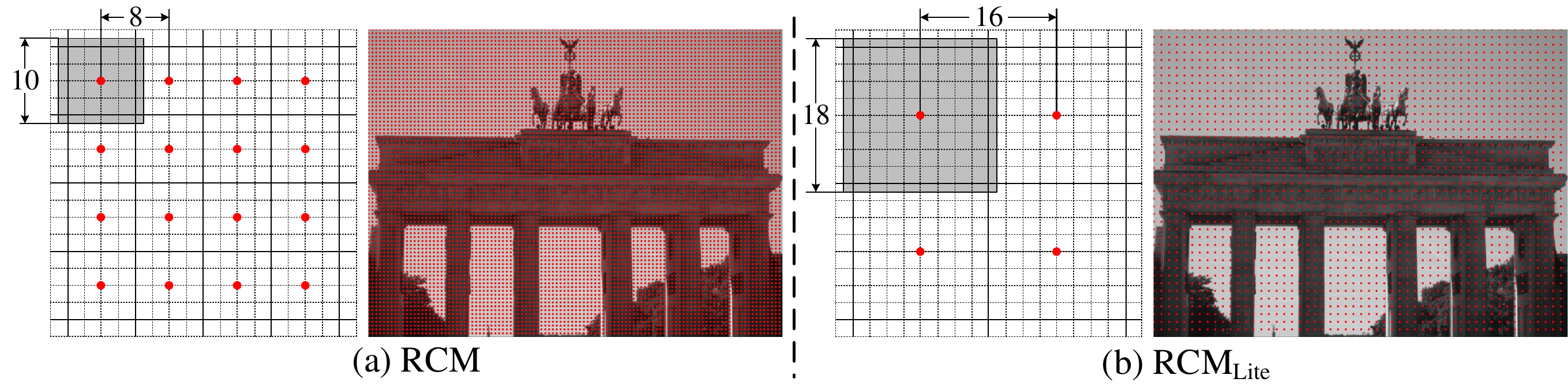}
	\centering
	 \caption{\textbf{Coarse grids (\textcolor{r}{red} points) and fine matching windows (\textcolor{gray}{gray} windows) in the target image.}}
	 \label{fig:resolution_window}
\end{figure}

\begin{figure*}[t]
	\includegraphics[width=0.99\linewidth]{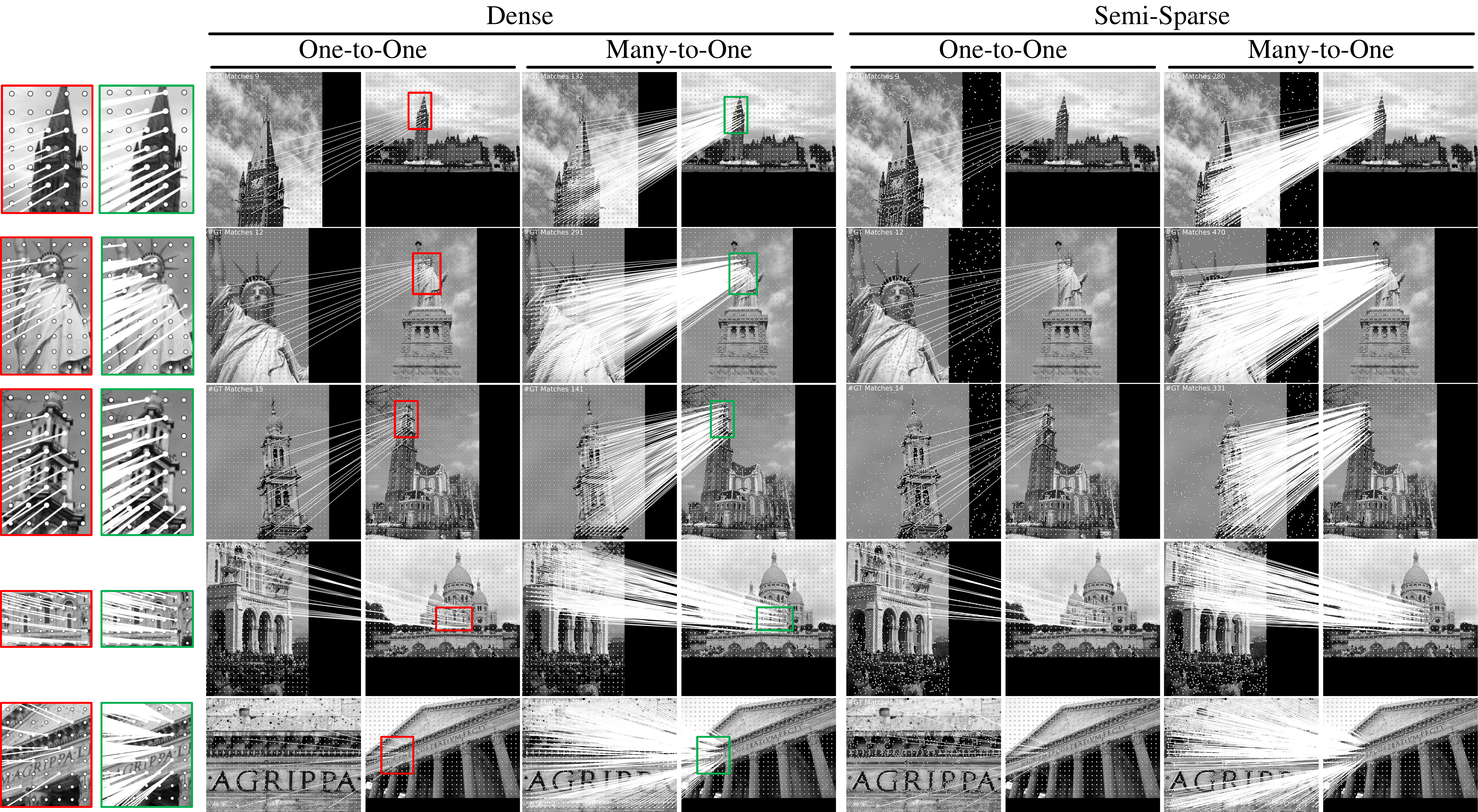}
	\centering
	 \caption{\textbf{Qualitative comparison of coarse ground-truth matches between ono-to-one and many-to-one matching.}
	 Many-to-one matching (\textcolor{g}{green} box) resolves the problem of matching conflicts in the target image, resulting in a greater number of ground-truth matches compared to one-to-one matching (\textcolor{r}{red} box).
	 Note that we only show the ground-truth for the coarse matching stage.}
	 \label{fig:gt_match_conflict}
\end{figure*}

\begin{figure*}[t]
	\includegraphics[width=0.99\linewidth]{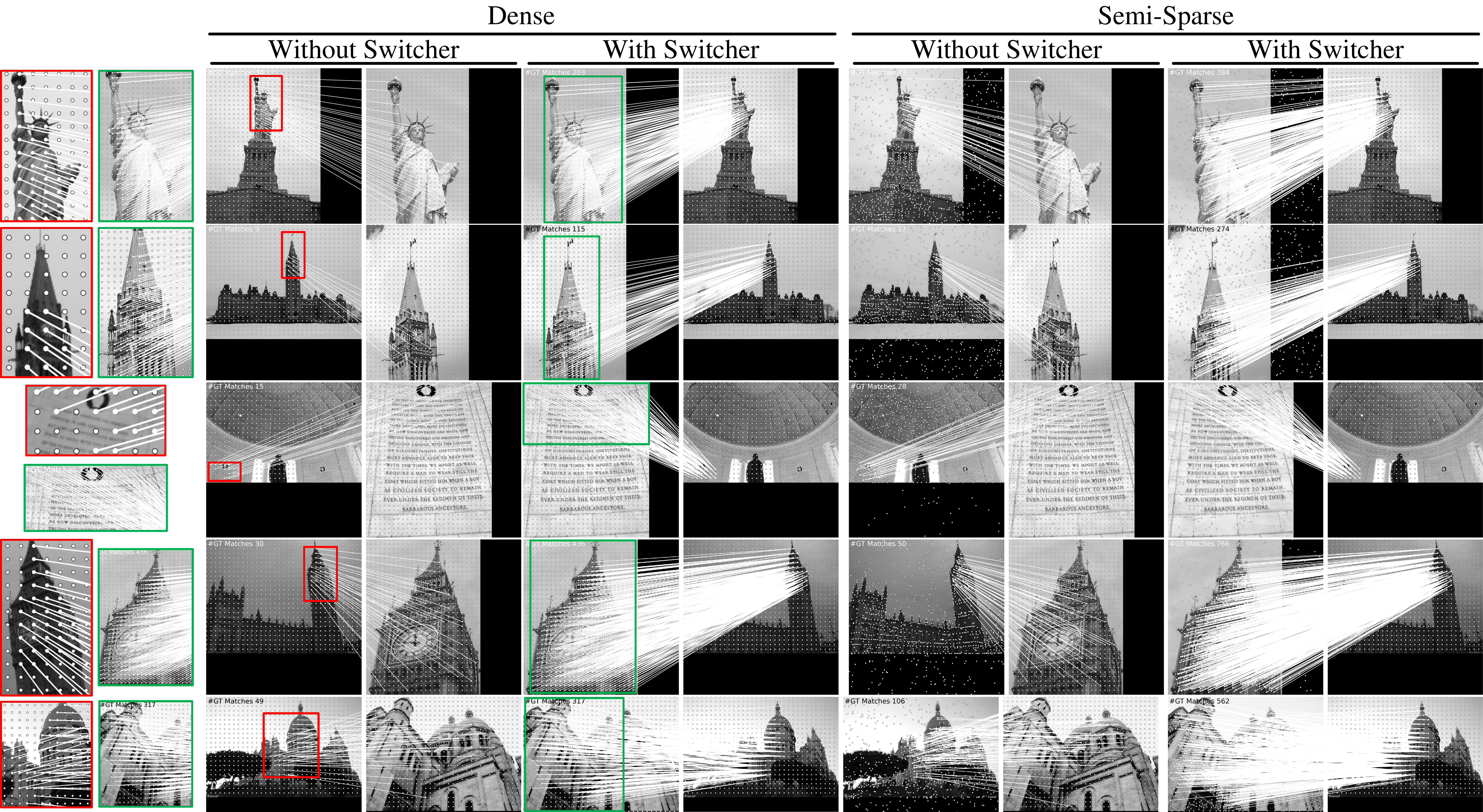}
	\centering
	 \caption{\textbf{Qualitative comparison of coarse ground-truth matches with and without switcher.}
	 The switcher resolves the problem of a shortage of matchable points (\textcolor{r}{red} box) in the source image, 
	 acquiring a significantly greater number of matchable points (\textcolor{g}{green} box) through the strategic switching of the two images.
	 }
	 \label{fig:gt_match_matchable}
\end{figure*}

\begin{figure*}[t]
	\includegraphics[width=0.99\linewidth]{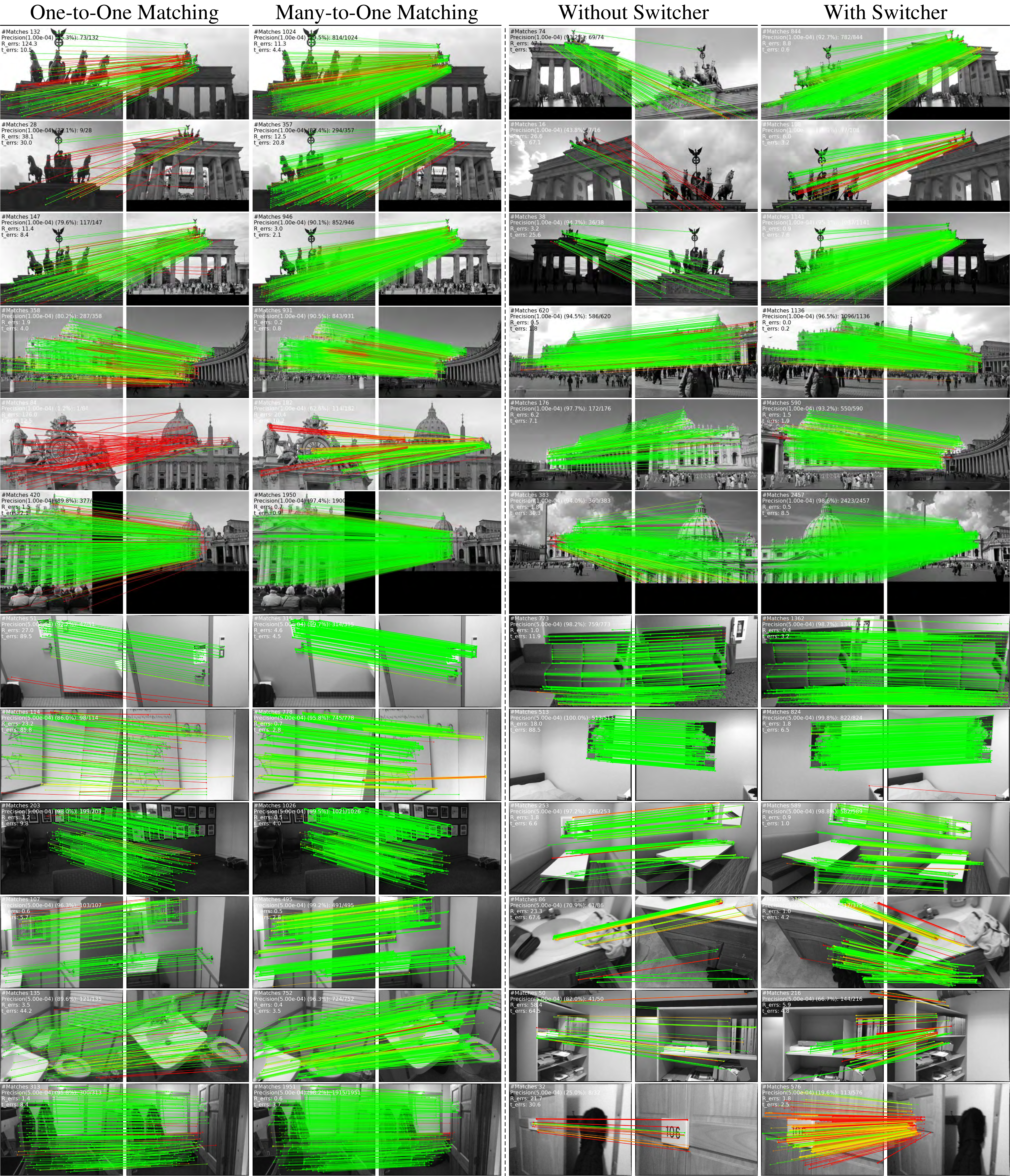}
	\centering
	 \caption{\textbf{Qualitatively results of many-to-one matching and switcher.}
	 }
	 \label{fig:m2o_switcher_viz_supp}
\end{figure*}

\begin{figure*}[t]
	\includegraphics[width=0.99\linewidth]{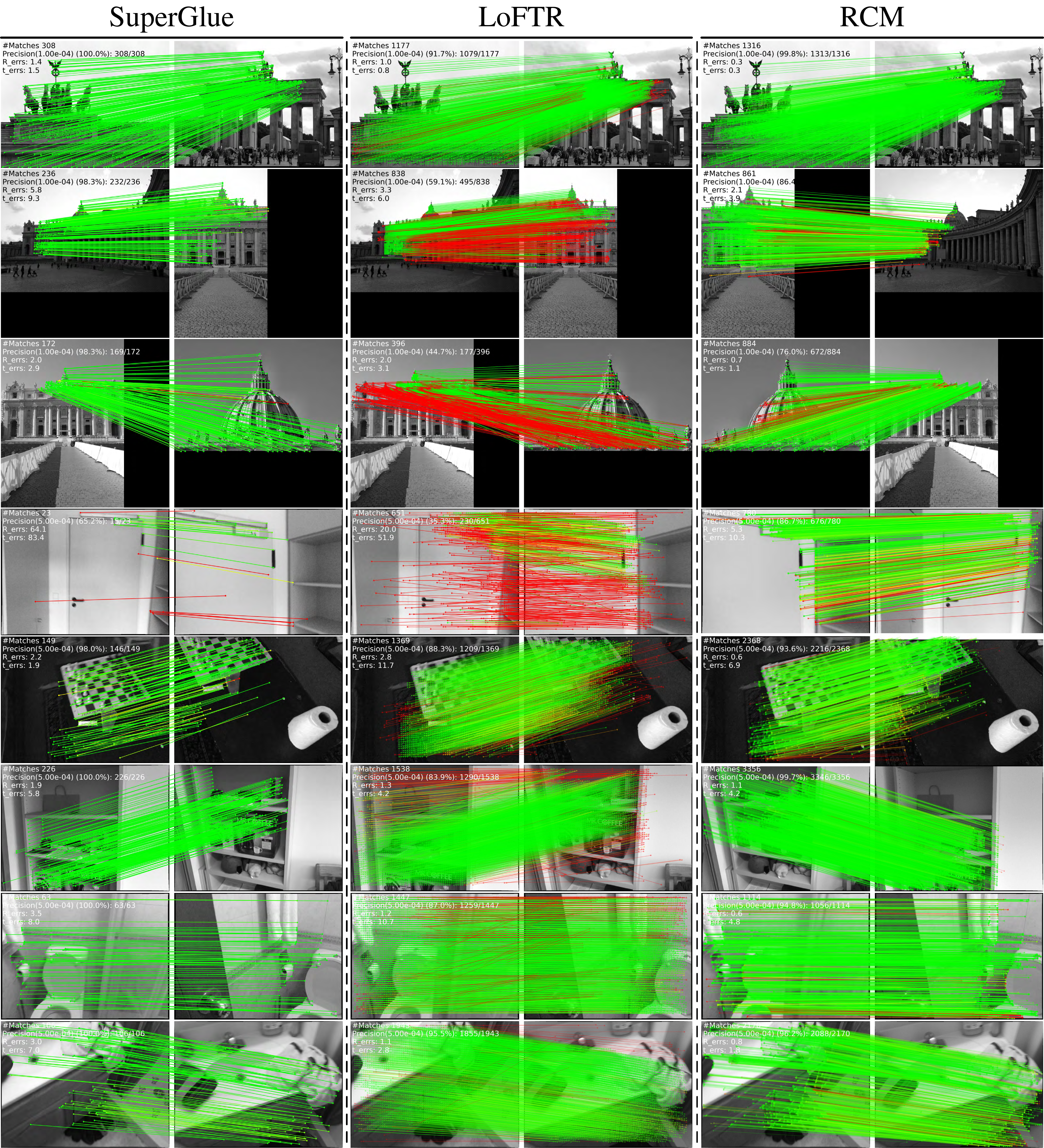}
	\centering
	 \caption{\textbf{Qualitative comparison in outdoor scenes.}
	 The semi-sparse matching method RCM consistently generates superior matches in both qualitative and quantitative aspects.}
	 \label{fig:viz_supp2}
\end{figure*}

\begin{figure*}[t]
	\includegraphics[width=0.99\linewidth]{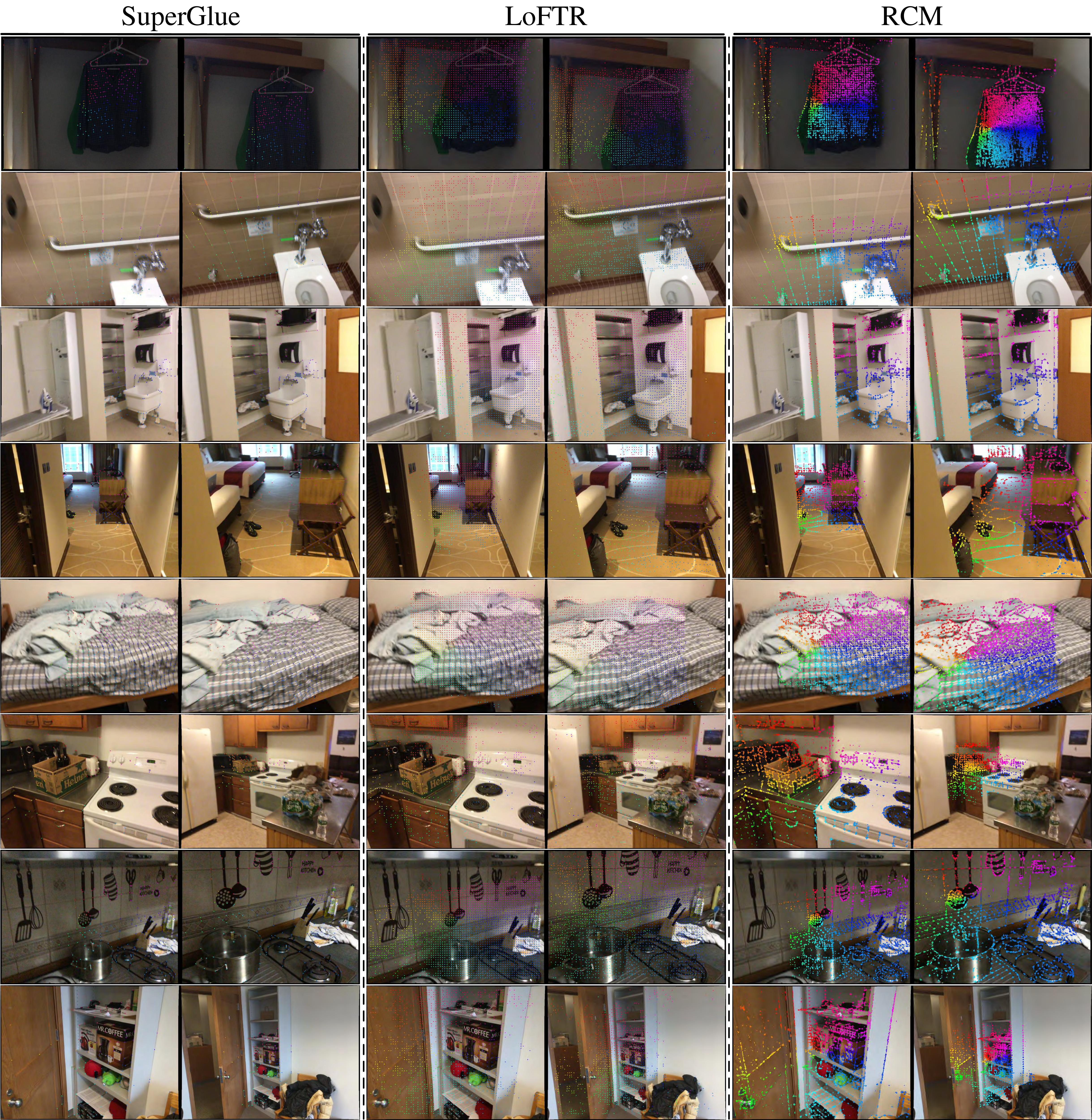}
	\centering
	 \caption{\textbf{More qualitative comparisons in indoor scenes.} The matched features are visualized as the same color.}
	 \label{fig:viz_wheel_in_supp}
\end{figure*}

\noindent\textbf{Coarse Feature Resolution and Fine Matching Window.}
As illustrated in \cref{fig:resolution_window}, 
RCM and RCM$_\mathrm{Lite}$ utilize coarse feature maps at resolutions of $1/8$ and $1/16$, respectively.
Consequently, each coarse feature corresponds to an $8\times8$ pixel patch for RCM and a $16\times16$ pixel patch for RCM$_\mathrm{Lite}$.
To ensure complete coverage of the coarse feature patch by the fine matching window,
window sizes of $10\times10$ and $18\times18$ pixels are designed for RCM and RCM$_{\mathrm{Lite}}$, respectively. 
Given that the fine feature map is at $1/2$ image resolution, the fine matching window sizes of RCM and RCM$_{\mathrm{Lite}}$ are $w=5$ and $w=9$, as discussed in the main text.

\noindent\textbf{Switcher.}
Initially, both sets of features undergo down-sampling to $(H,W)=(20,20)$ using adaptive average pooling to optimize subsequent computations. 
The correlation map $C\in\mathbb{R}^{20 \times 20 \times 20 \times 20}$ is then computed by the inner product, 
capturing the similarity of individual patches between the two images. 
The correlation map is reshaped into $\hat{C}\in\mathbb{R}^{20 \times 20 \times 400}$ and processed by a lightweight CNN,
which extracts features through two Conv-BN-ReLu-MaxPool layers.
We subsequently reduce the spatial dimension to $1$ with adaptive average pooling and the channel dimension to $2$ with linear layer. 
$\mathrm{Softmax}$ is applied to compute the switching confidence $VS$, triggering feature switching when $VS>1/2$.

\newpage

\begin{figure*}[t]
	\includegraphics[width=0.99\linewidth]{graph/viz_wheel_out_supp_s.pdf}
	\centering
	 \caption{\textbf{More qualitative comparisons in outdoor scenes.} The matched features are visualized as the same color.}
	 \label{fig:viz_wheel_out_supp}
\end{figure*}

\begin{figure*}[t]
	\includegraphics[width=0.99\linewidth]{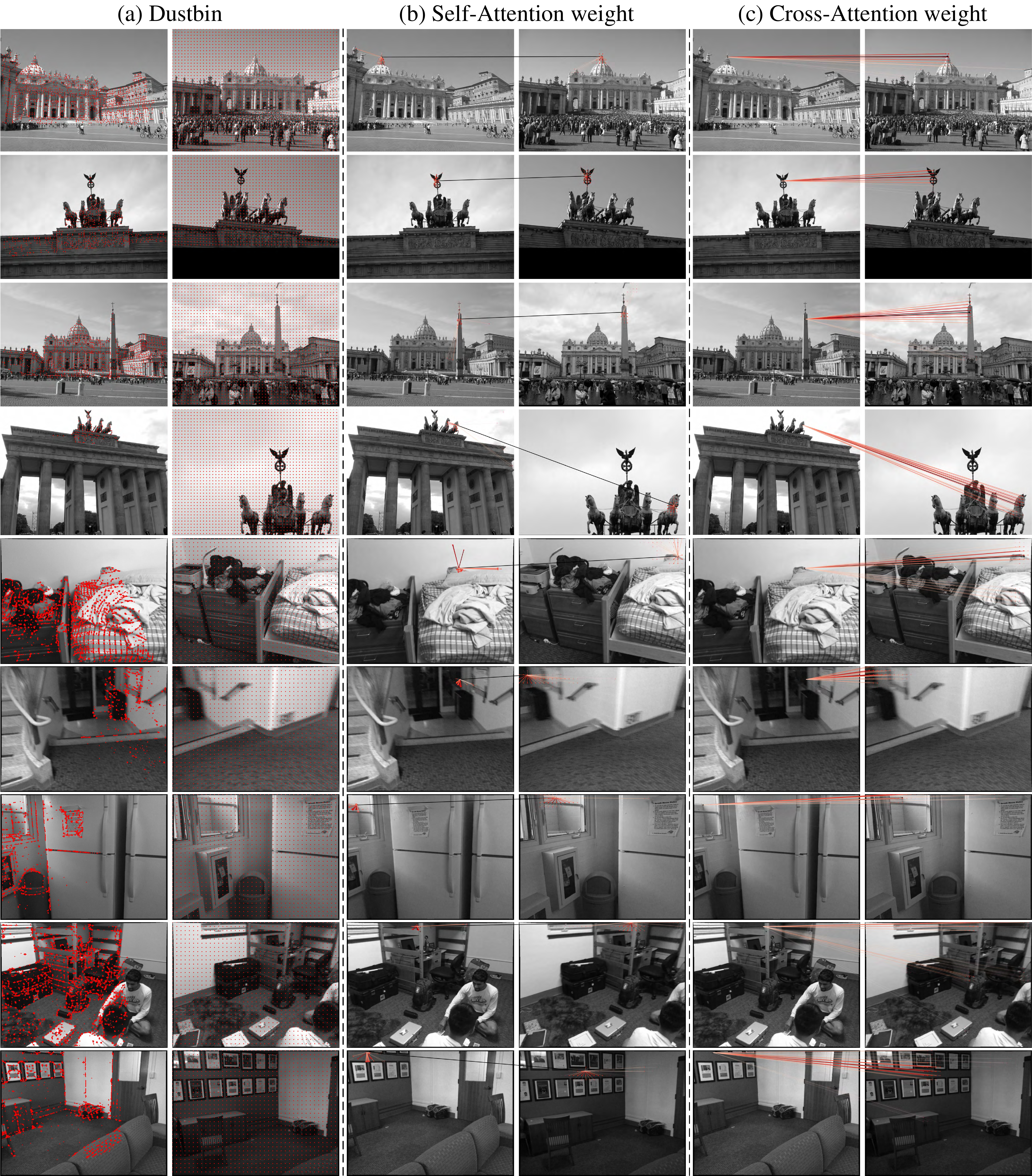}
	\centering
	 \caption{\textbf{Visualizations of the dustbin and attention weights.}}
	 \label{fig:viz_dustbin_attnn_supp}
\end{figure*}

\begin{figure*}[h]
	\includegraphics[width=0.99\linewidth]{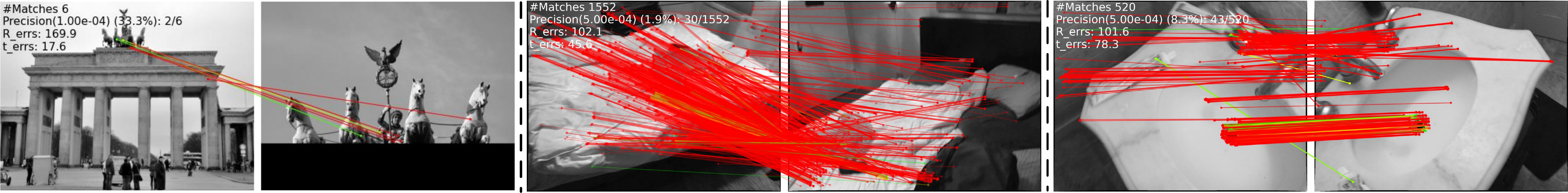}
	\centering
	 \caption{\textbf{Failure cases.}}
	 \label{fig:fail_supp}
\end{figure*}

\begin{figure*}[h]
	\includegraphics[width=0.99\linewidth]{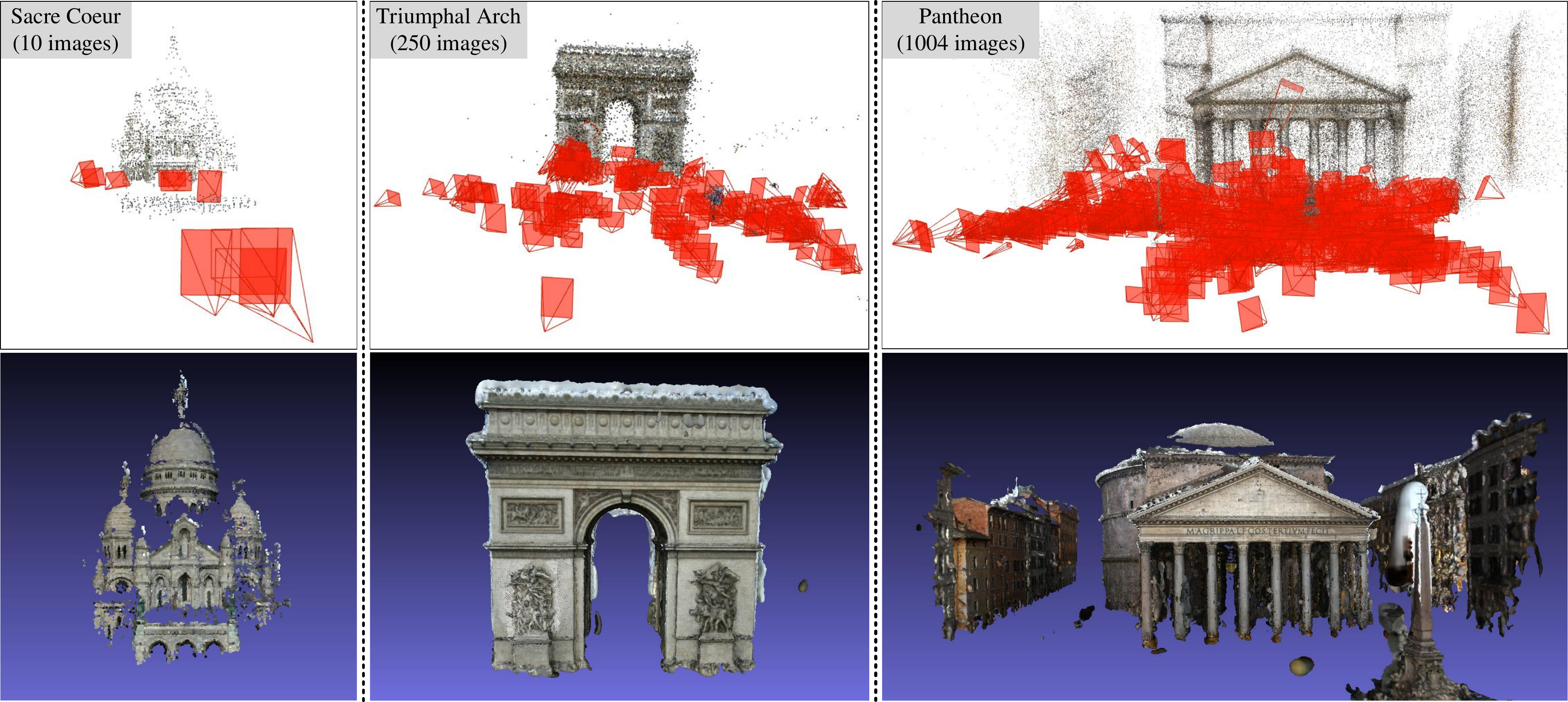}
	\centering
	 \caption{\textbf{3D reconstruction results based on RCM.}}
	 \label{fig:viz_reconstruction_supp}
\end{figure*}

%
%

\end{document}